\documentclass[conference]{IEEEtran}
\IEEEoverridecommandlockouts
\usepackage{cite}
\usepackage{amsmath,amssymb,amsfonts}
\usepackage{algorithm}
\usepackage{algorithmic}
\usepackage{graphicx}
\usepackage{textcomp}
\usepackage{xcolor}
\usepackage{diagbox}
\usepackage{hyperref}
\usepackage{booktabs}
\usepackage[misc]{ifsym}

\usepackage{mwe}
\usepackage{subcaption}
\usepackage{pgfplots}
\usepackage{filecontents}
\usepgfplotslibrary{groupplots, fillbetween, statistics}
\pgfplotsset{
/pgfplots/boxplot legend/.style={
   legend image code/.code={
     \draw [|-|,##1] (0,2mm) --
         node[rectangle,minimum size=2.5mm,draw,fill,##1]{}
        (0,7mm);
    }
  }
}
\pgfplotsset{width=7cm,compat=1.8,/pgf/bar width=5pt 
}
\usepackage{mathtools}
\usepackage{amssymb}
\usepackage{multirow}
\usepackage{float}
\usepackage{enumitem}

\newcommand{\subfour}[1]{\vspace*{3mm}{\noindent\bf #1}}

\def\BibTeX{{\rm B\kern-.05em{\sc i\kern-.025em b}\kern-.08em
    T\kern-.1667em\lower.7ex\hbox{E}\kern-.125emX}}
\begin{document}

\title{Margin-bounded Confidence Scores for Out-of-Distribution Detection\\
{\footnotesize \textsuperscript{*}Accepted as a regular paper at ICDM 2024}
}
\author{\IEEEauthorblockN{Lakpa D. Tamang,
Mohamed Reda Bouadjenek, Richard Dazeley, and
Sunil Aryal}
\IEEEauthorblockA{Department of Science, Engineering, and Built Environment, School of Information Technology,
Deakin University\\
Geelong, Victoria, Australia\\
l.tamang@research.deakin.edu.au,
\{reda.bouadjenek,
richard.dazeley,
sunil.aryal\}@deakin.edu.au}}

\maketitle

\begin{abstract}
In many critical Machine Learning applications, such as autonomous driving and medical image diagnosis, the detection of out-of-distribution (OOD) samples is as crucial as accurately classifying in-distribution (ID) inputs.
Recently \textit{Outlier Exposure} (OE) based methods have shown promising results in detecting OOD inputs via model fine-tuning with auxiliary outlier data. However, most of the previous OE-based approaches emphasize more on synthesizing extra outlier samples or introducing regularization to diversify OOD sample space, which is rather unquantifiable in practice. In this work, we propose a novel and straightforward method called \textbf{Ma}rgin bounded \textbf{C}onfidence \textbf{S}cores (MaCS) to address the nontrivial OOD detection problem by enlarging the disparity between ID and OOD scores, which in turn makes the decision boundary more compact facilitating effective segregation with a simple threshold. Specifically, we augment the learning objective of an OE regularized classifier with a supplementary constraint, which penalizes high confidence scores for OOD inputs compared to that of ID and significantly enhances the OOD detection performance while maintaining the ID classification accuracy. Extensive experiments on various benchmark datasets for image classification tasks demonstrate the effectiveness of the proposed method by significantly outperforming state-of-the-art (S.O.T.A) methods on various benchmarking metrics. The code is publicly available at \url{https://github.com/lakpa-tamang9/margin_ood}
\end{abstract}


\section{Introduction}
\label{sec:intro}
Machine learning-based systems in critical applications such as autonomous driving and medical image diagnosis should equally prioritize accurately classifying in-distribution (ID) inputs and detecting out-of-distribution (OOD) samples, which are also referred to as anomalies or novelties.
This issue may arise because real-world data are dynamic in nature, where distribution shifts frequently occur owing to the emergence of new classes, leading to significant differences in the posterior probabilities of input and labels \cite{yang2021generalized}. 
Hence, a classification system must avoid classifying objects from unknown classes to establish user trust.

\begin{figure}[t]
\centering
\includegraphics[width = 0.45\textwidth]{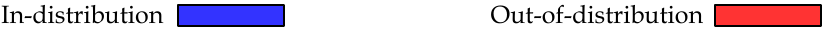}
\begin{tikzpicture}
  \begin{axis}
    [
    boxplot/draw direction=y,
    ytick = {0.0, 0.2, 0.4, 0.6, 0.8, 1.0},
    ylabel = Confidence Score,
    height = 4cm,
    width = 9cm,
    xtick = {1,...,11},
    xticklabels={,MSP \cite{hendrycks2016baseline},,,,OE \cite{hendrycks2018deep},,,,MaCS},
    grid = minor,
    boxplot/variable width,
    boxplot/box extend=0.7,
    legend style={
    legend pos = outer north east, 
    /tikz/every even column/.append style={column sep=0.1cm},
    /tikz/every odd column/.append style={column sep=0.1cm, row sep=0.1cm},
    /tikz/every node/.append style={align=center, yshift=0.05cm} 
    },
    boxplot legend,
    cycle list={{draw=black, fill=blue!80},{draw=black, fill=red!80},{draw=black, fill=blue!80},{draw=black, fill=red!80}}
    ]
    \addplot+[
    boxplot prepared={draw position =9,
      median=0.8850863873958588,
      upper quartile=0.51356540620327,
      lower quartile=0.9900374263525009,
      upper whisker=1.0,
      lower whisker=0.046664975583553314
    },
    ] coordinates {};
    \addplot+[
    boxplot prepared={draw position = 10.2,
      median=0.21252813190221786,
      upper quartile=0.11789200082421303,
      lower quartile=0.42142052203416824,
      upper whisker=0.8747148513793945,
      lower whisker=0.02586529031395912
    },
    ] coordinates {};
    \addplot+[
    boxplot prepared={draw position = 5,
      median=0.9502650201320648,
      upper quartile=0.628040537238121,
      lower quartile=0.9975472688674927,
      upper whisker=1.0,
      lower whisker=0.09451377391815186
    },
    ] coordinates {};
    \addplot+[
    boxplot prepared={draw position = 6.2,
      median=0.5064583420753479,
      upper quartile=0.20724257454276085,
      lower quartile=0.9086686968803406,
      upper whisker=1.0,
      lower whisker=0.040033742785453796
    },
    ] coordinates {};
    \addplot+[
    boxplot prepared={draw position = 1,
      median=0.950387716293335,
      upper quartile=0.6599675714969635,
      lower quartile=0.9984955340623856,
      upper whisker=0.9999985694885254,
      lower whisker=0.18760256469249725
    },
    ] coordinates {};
    \addplot+[
    boxplot prepared={draw position = 2.2,
      median=0.7171927690505981,
      upper quartile=0.502552293241024,
      lower quartile=0.9253356605768204,
      upper whisker=0.9999524354934692,
      lower whisker=0.12953081727027893
    },
    ] coordinates {};
  \end{axis}
\end{tikzpicture}

\caption{Confidence scores of models trained using CIFAR-100 on test data from CIFAR-100 (ID samples) and iSUN \cite{xu2015turkergaze} (OOD samples).
}
\label{fig:figure_1}
\end{figure}
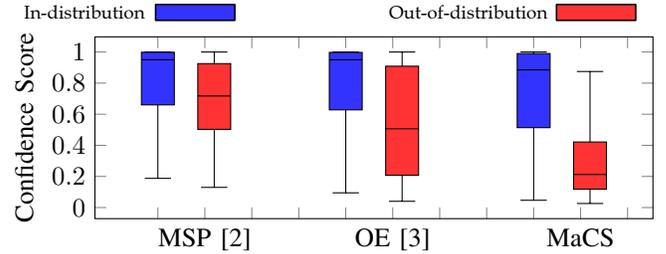

OOD detection is a classic yet essential ML problem that aims to resolve the fundamental issue of models being overconfident in classifying samples from different semantic distributions \cite{yang2021generalized}.
Hence, numerous approaches have been proposed to solve this task~
\cite{scholkopf1999support,chandola2009anomaly,noble2003graph,scheirer2012toward,scheirer2014probability,japkowicz1995novelty}, which typically rely on a post-hoc detection strategy, employing thresholds or other criteria to identify OOD samples.
Another technique that has attracted considerable attention is the Outlier Exposure (OE) method  \cite{hendrycks2018deep} that advocates the use of outliers to regularize the model and generate low confidence scores on unseen distributions.
To compare the confidence scores, i.e., the maximum values of the Softmax probabilities of ID and OOD samples for some of these techniques, we refer to Fig. \ref{fig:figure_1}.
Here, we train three image classification models -- Maximum Softmax Probability (MSP) \cite{hendrycks2016baseline}, OE \cite{hendrycks2018deep}, and our proposed method MaCS -- on the CIFAR-100 dataset. We use test images from CIFAR-100 as ID samples and images from the iSUN dataset \cite{xu2015turkergaze} as OOD samples. In the literature, these two datasets serve as popular benchmark datasets utilized for the OOD detection task; the former is primarily employed as ID, while the latter is used to represent OOD data.
We employ boxplots for visualization and score comparison, from which we observe the following:
First, the MSP method, a straightforward classification model that optimizes cross-entropy, exhibits overconfidence when applied to OOD samples as their scores overlap significantly with those of ID samples. Second, while OE generally helps to decrease the scores of OOD samples, the overlap between the scores of OOD and ID samples is still noteworthy. The reason for this is that outliers can occasionally produce confidence scores comparable to, or even higher than those of ID samples. As a result, OOD samples that lie in the decision boundary can be often falsely categorized as ID data, which poses a challenge in their clear separation.

Moreover, most OOD detection methods rely on sampling and synthesizing existing outliers \cite{chen2021atom}, \cite{zhu2024diversified}, introducing regularization through augmentations \cite{zhu2023openmix,zhang2023mixture}, and feature space maneuvering \cite{wang2023out}. While these approaches attain reasonable detection performance, they may often suffer from a phenomenon, which we refer to as ``score explosion'', where the confidence score for OOD samples exceeds that of ID samples as shown in Fig. \ref{fig:figure_1}.
To address this issue, this paper introduces a novel approach called Margin bounded Confidence Scores (MaCS). 
Leveraging the insight gained from score explosions, MaCS penalizes the model during training, encouraging it to learn discriminative features between ID data and representative outliers. 
By nullifying score explosions and assigning weights based on the margin difference between ID and OOD confidence scores, MaCS aims to reduce model uncertainty in distinguishing between the two. 
In Fig. \ref{fig:figure_1}, the last two boxplots illustrate the distribution of scores for OOD and ID samples under MaCS, where clearly OOD samples receive significantly lower confidence scores compared to ID samples.

The {\bf contributions} of this paper can be summarised as:
\begin{itemize}
    \item \textbf{Simple and Practical Solution: }We investigate an OOD detection problem under a practical research setting, utilizing the existing confidence scores of any OE regularized model: a completely different approach compared to conventional outlier synthesis techniques whose objective is establishing heterogeneity of OOD sample space that cannot be quantitatively measured in practice. 
  \item \textbf{Learning in Synergy: }We propose a novel and straightforward method called \textbf{Ma}rgin bounded \textbf{C}onfidence \textbf{S}cores (MaCS) that work together with OE under a unified algorithm: a supplementary constraint is put forward to the training objective of the OE method to enhance the OOD detection robustness of a classification model.
  \item \textbf{Effectiveness: }We conduct comprehensive experiments utilizing established benchmark ID and OOD image classification datasets. Our findings reveal significant enhancements over several state-of-the-art (S.O.T.A) methods across various detection metrics. Furthermore, we validate our method by performing several ablation studies and prove it to be highly effective in achieving reliable detection performance under different networks, and datasets.
\end{itemize}

\section{Related Works}
There is a substantial body of research related to OOD detection techniques. 
Below, we review the major works related to post-hoc OOD detection and OOD detection by using auxiliary outliers.

\subfour{Post-hoc OOD Detection: }
Post-hoc OOD detection techniques have the advantage of being easy to use without modifying the training procedure and objective of the model. In this regard, various scoring functions have been proposed to better utilize the high level semantic information of penultimate layers. A MaxLogit technique \cite{hendrycks2022scaling} uses the maximum value of logits instead of softmax probabilities to enhance the detection performance. In the following works, \cite{jung2021standardized} used standarized value of maximum logit scores to align different distributions, and \cite{zhang2023decoupling} decoupled the maximum logits value for flexibility to balance MaxCosine and MaxNorm. Similarly, ODIN \cite{liang2017enhancing} and Generalized ODIN \cite{hsu2020generalized} proposed the decomposition of confidence scores and modified input pre-processing methods to enhance detection performance. Additionally, ReAct \cite{sun2021react} used activation rectification during the test time for stronger separation of ID and OOD data and DICE \cite{sun2022dice} used weight ranking to select the most salient weights to derive the OOD detection output.

\subfour{OOD Detection by Using Auxiliary Datasets: }
Generating outliers or auxiliary OOD examples is essential to improve the robustness and generalization capabilities of a model \cite{chen2021atom}. The goal is to expose the model to a wider range of data scenarios beyond what is available in the training set. In literature, OOD detection has been realized by producing synthetic outliers using methods such as data augmentation \cite{pinto2022using,yun2019cutmix,zhu2023openmix}, and adversarial example generation \cite{lee2018training,mohseni2020self,papadopoulos2021outlier,zheng2024out}. One such method, Energy OOD \cite{liu2020energy}, uses energy scores instead of softmax scores because they are more aligned with the probability density of the inputs and are less prone to overconfidence. Another related study, GEM \cite{morteza2022provable}, models the feature space as a class-conditional multivariate Gaussian distribution. MixOE \cite{zhang2023mixture} and MiM \cite{choi2023towards} used MixUp regularizers to mix ID data with auxiliary outliers, with the former being in complex fine-grained scenarios.
Motivated by the recent achievements of auxiliary outliers based approaches, our objective is to harness it's potential for OOD detection. Unlike other methods that depend partially or entirely on data augmentation-based regularization \cite{zhang2023mixture,choi2023towards} and intricate outlier synthesis/sampling techniques \cite{wang2023out,chen2021atom}, we present a less sophisticated method that relies on the confidence scores of a model while using eminent outlier datasets.

\section{Background}
\label{sec:notations}

\subsection{Notation and Problem Definition}

We consider a training dataset independently and identically distributed (\textit{i.i.d}) data drawn from ID, $\mathcal{D}_{in}= \{(\textbf{x}^{(1)},y^{(1)}), (\textbf{x}^{(2)},y^{(2)}),$ $\cdots, (\textbf{x}^{(k)},y^{(k)})\}$ with $k$ instances, where each $\textbf{x}^{(i)}\in \mathcal{X} \subseteq \mathbb{R}^n$ is an $n$-dimensional input feature vector of the instance $i$, and $y^{(i)}\in \mathcal{Y}=\{y_1,y_2,\cdots,y_c\}$ represents its class. 
Similarly, during test phase, we evaluate the OOD detection capability using examples drawn from the OOD sample space $\mathcal{D}_{out}= \{\textbf{x}^{(1)}, \textbf{x}^{(2)},$ $\cdots, \textbf{x}^{(k)}\}$.
Also, following the convention in \cite{hendrycks2018deep}, we introduce auxiliary outlier data as $\mathcal{D}_{out}^{OE}$ such that $\mathcal{D}_{out}^{OE} \cap \mathcal{D}_{out} \cap \mathcal{D}_{in} = \phi$.

The goal is then to learn a mapping function $f:\mathcal{X}\rightarrow \mathbb{R}^{c}$ trained using $\mathcal{D}_{in}\cup\mathcal{D}_{out}^{OE}$, which assigns to each feature vector $\textbf{x}^{(i)}\in \mathcal{D}_{in}$  its correct class $y^{(i)}$, while avoiding classifying  instances $\textbf{x}^{(i)}\in \mathcal{D}_{out}^{OE}$.

\subsection{Outlier Exposure}
Outlier Exposure (OE), an auxiliary outlier based OOD detection method \cite{hendrycks2018deep} is the baseline that we refer to in our study. It is a regularization technique that involves learning from additional datasets containing outliers or OOD samples with low confidence predictions along with standard training data. The goal is to expose the network to diverse OOD examples during training, so that the model learns a more conservative concept of the ID data to distinguish them from their OOD counterparts. To achieve this, OE uses an auxiliary dataset of outliers $\mathcal{D}_{out}^{OE}$ that is entirely separate from the OOD test data $\mathcal{D}_{out}$. 
Hence, OE is trained by optimizing the following objective:

\begin{equation} \label{eq:1}
\mathcal{L}_{OE}=\mathbb{E}_{(\textbf{x},y)\sim\mathcal{D}_{in}}\left[\mathcal{L}(f(\textbf{x}),y)\right] + \lambda_1\mathbb{E}_{\textbf{x}\sim\mathcal{D}_{out}^{OE}}[\mathcal{L}(f(\textbf{x}),\mathcal{U})]
\end{equation}

\noindent where  $\mathcal{L}$ is the cross-entropy loss, $\mathcal{U}\in\mathbb{R}^{k}$ represents a uniform distribution over $c$ classes, and $\lambda_1$ is the hyper-parameter for balancing both objectives.

\subsection{Scoring Function}
We adopt MSP as a method for detecting OOD samples, which operates using a threshold. 
MSP retains the maximum posterior probability (or confidence scores) over  softmax probabilities of a network \cite{hendrycks2016baseline}. 
Thus, if $\textbf{s}(\textbf{x})=\left\{s_1, s_2, ..., s_c\right\}$ denotes the confidence scores across $c$ classes, the MSP is represented by $\max(\textbf{s}(\textbf{x}))$. 
In essence, by comparing this value with a predetermined threshold $\tau \in \left\{0, 1\right\}$, we can classify a given test input as either ID or OOD.

\begin{equation} \label{eq:2}
  g(\textbf{x})=\begin{cases}
    \mathcal{ID}, & \text{if $\max(\textbf{s}(\textbf{x}))\geq\tau$}.\\
    \mathcal{OOD}, & \text{otherwise}.
  \end{cases}
\end{equation}

\section{Proposed Method: Margin bounded Confidence Scores (MaCS) }
\label{sec:proposed}
\begin{figure*}[ht!]
    \centering
    \includegraphics[scale=0.6]{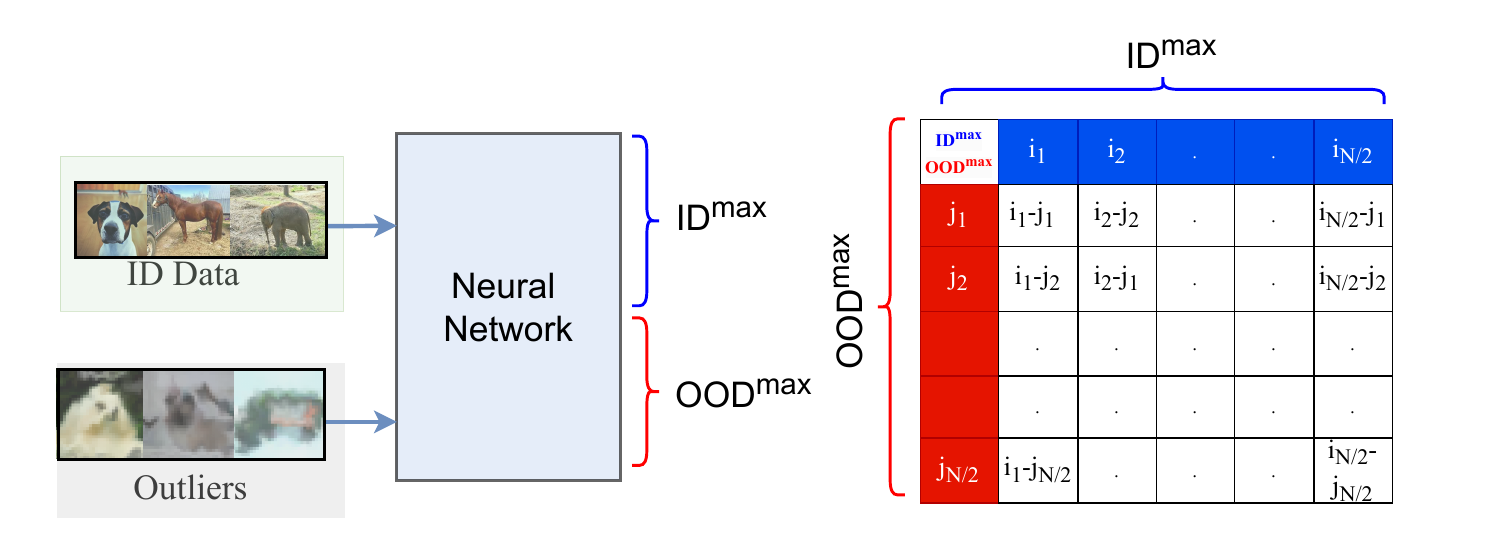}
    \caption{Schematic overview of MaCS where the maximum confidence scores of inputs from $\mathcal{D}_{in}$ and $\mathcal{D}_{out}^{OE}$ are extracted from the output layer of neural network followed by element-wise difference computation between $ID^{max}$ and $OOD^{max}$.}
    \label{fig:figure_2}
\end{figure*}

In this section, we introduce the MaCS framework. 
Initially, we augment Equation \eqref{eq:1} with an additional loss component aimed at promoting a distinct separation between ID and OOD samples. 
Fig. \ref{fig:figure_2} illustrates our approach, wherein we compute $\max(\textbf{s}(\textbf{x}))$ for inputs from both $\mathcal{D}_{in}$ and $\mathcal{D}_{out}^{OE}$, followed by subtracting the former from the latter. 
We refer to this operation as Maximum Confidence Difference (MCD), which is elaborated on in Section \ref{sec:mcd}. 
Subsequently, we address score explosions, where the confidence score of the outlier exceeds that of the ID input. 
Finally, we constrain these score differences within a specified margin value. 
Further details regarding margin-based weighting are provided in Section \ref{sec:boundedmargin}.

\subsection{Maximum Confidence Difference (MCD) and Penalty}
\label{sec:mcd}
We consider an input to the model, with equiproportionate samples from $\mathcal{D}_{in}$ and $\mathcal{D}_{out}^{OE}$ such that a batch 
$\mathcal{B}=\left\{\textbf{x}_i\right\}^{2N}_{i=1}$ has $N$ samples from $\mathcal{D}_{in}$ and $N$ samples from $\mathcal{D}_{out}^{OE}$ with the batch size of $2N$.
We obtain confidence scores for $\mathcal{B}$ denoted as  $\mathbf{S}_{\mathcal{B}}=\left\{\mathbf{s}(\textbf{x}_i)\right\}_{i=1}^{2N}$.
Next, we compute the maximum confidence score for each instance $\textbf{x}_i \in \mathcal{B}$ as $\max(\mathbf{s}(\textbf{x}_i))$. We denote these maximum scores as $ID^{max}$ and $OOD^{max}$ for inputs from $\mathcal{D}_{in}$ and $\mathcal{D}_{out}^{OE}$, respectively. Note that both $ID^{max}$ and $OOD^{max}$ are $N$-dimensional vectors.
Intuitively, the $\max(\mathbf{s}(\textbf{x}_i))$ represents the notion of confidence of the model to categorize $\textbf{x}_i$ into one of $c$ classes. 
Subsequently, for each element of $ID^{max}$ we compute the difference between every element of $OOD^{max}$. For instance, if there  (see Fig. \ref{fig:figure_2} for graphical illustration). 
We do this to ensure that every OOD input whose $\max(\mathbf{s}(\textbf{x}_i))$ is larger than that of the ID is captured. 
Following that, we employ ReLU to penalize these occurrences by setting the negatives to zero, while retaining only the positives. 
Finally, the Maximum Confidence Difference (MCD) of batch $\mathcal{B}$ is estimated as:

\begin{equation}
\mathcal{MCD}(\mathcal{B}) = \frac{1}{N} \sum_{i=1}^{N} \sum_{j=1}^{N} \max(0, ID^{max}_i - OOD^{max}_{j})^2
\label{eq:3}
\end{equation}

\subsection{Bounded Margin}
\label{sec:boundedmargin}
Furthermore, we bound the overall MCD term to be within a specified range to distinctly dispel the ID and OOD data thus subtracting it from the margin value. The idea is that for a correctly-classified ID sample, the model should not only be confident about it being correctly classified but also confident that it is not an OOD. Thus, we aim to give considerable attention to the OOD samples by assigning a weight $\mathcal{W}_{MaCS}$. 
We follow a similar idea of the weighting approach \cite{yue2017imbalanced,fernando2021dynamically}, which attempts to solve class imbalance problem in classification tasks. 
Similar to how weights are administered to make the model more sensitive towards under-sampled classes, we attempt to assign weights to rectify the exploded scores. In particular, we typically assign weights based on the occurrence of score explosions, instead of class memberships \cite{anand2010approach,zong2013weighted}.
This phenomenon is crucial for OOD detection, where failing to detect an OOD sample is considered as severe as misclassifying an ID sample. 
With this, we define a more tailored weighting strategy that explicitly addresses the nature of the error, which is OOD scores exceeding ID scores rather than focusing on the under-represented classes. Mathematically, we administer $\mathcal{W}_{MaCS}$ as follows: 

\begin{equation}
    \mathcal{W}_{MaCS}=\max(0, m - \mathcal{MCD}(\mathcal{B}))
\label{eq:4}
\end{equation}

\noindent where $m$ is the margin that enforces the minimum difference between the ID and OOD output. 
The best value of $m$ is determined empirically and as explained in Section \ref{sec:margin}. 
To put this into perspective, if MCD goes to zero, we replace it with $\mathcal{W}_{MaCS}$, which relates to weight assignment for score explosions. As a supplement to the training objective of OE, we combine the term in \eqref{eq:4} with \eqref{eq:1} resulting in our final training objective for the whole dataset with $B$ batches as follows:

\begin{equation} \label{eq:5}
\mathcal{L}_{final}=\sum_{i = 1}^{B}(\mathcal{L}_{OE}^{(i)}+ \lambda_2\mathcal{W}_{MaCS}^{(i)})
\end{equation}

where, $\lambda_2$ is a hyperparameter for balancing the effect of weighted margin on $\mathcal{L}_{OE}$. We summarize the whole procedure of fine-tuning MaCS as a pseudo-code in Algorithm \ref{algorithm_1}. 

\begin{algorithm}[ht!]
    \caption{Fine-tuning Margin Bounded Confidence Scores}\label{algorithm_1}
    \begin{algorithmic}[0]
    \STATE \textbf{Input:} $\mathcal{D}_{in}$, $\mathcal{D}_{out}^{OE}$, pre-trained model $f$, hyperparameters $\theta$, epochs $T$, and margin $m$;
    \STATE \textbf{Output:} finetuned model $f^*$ with $\theta^*$, and $m^*$;
    \end{algorithmic}
    \begin{algorithmic}[1]
    \FOR{$m$ = 0.0 to 0.9 with step-size of 0.1} 
        \FOR{epoch = 1 to $T$}
            \FOR{batch = 1 to $B$}
                \STATE Select a batch $\mathcal{B}=2N$, with $N$ outliers, and $N$ ID inputs from $\mathcal{D}_{out}^{OE}$, and  $\mathcal{D}_{in}$ respectively;
                \STATE Concatenate sampled data from $\mathcal{D}_{in}$, and $\mathcal{D}_{out}^{OE}$ to create new input data, $\mathbf{x}$; 
                \STATE Calculate $f(\mathbf{x}; \theta)$, to get confidence scores;
                \STATE Compute maximum confidence score for each input with MSP;
                \STATE Compute MCD using \eqref{eq:3};
                \STATE Compute $\mathcal{W}_{MACS}$ using \eqref{eq:4}
                \STATE Compute overall loss using \eqref{eq:5} 
            \ENDFOR
        \ENDFOR
    \ENDFOR
    \end{algorithmic}
\end{algorithm}

\section{Experiments and Results}
\label{sec:experiments}
This section outlines our experimental setup for conducting methodological evaluation, which include details regarding the benchmark datasets, baselines, and metrics utilized in our analysis.

\subsection{Datasets}
\label{sec:datasets}
We categorize our data into three types: ID, outlier, and OOD datasets. 
The ID and outlier datasets are explicitly designated for training or fine-tuning purposes, while the OOD datasets are reserved for testing scenarios only.

\subsubsection{ID Datasets}
\label{sec:id_dataset}
Our experiments are performed on four different image datasets:
(1) \textbf{CIFAR-10}~\cite{krizhevsky2009learning}: A small image classification dataset with 10 classes;
(2) \textbf{CIFAR-100}~\cite{krizhevsky2009learning}: A medium-scale image classification dataset with 100 classes;
(3) \textbf{SVHN}~\cite{netzer2011reading}: A small-scale image dataset with 10 classes, consisting of digits from 0 to 9;
and (4) \textbf{Imagenet-32 }~\cite{chrabaszcz2017downsampled}: A down-sampled version of the original Imagenet-1k \cite{russakovsky2015imagenet}, which is considered a large-scale dataset due to its 1,000 classes.
Note that our training, validation, and test data follow the standard splits provided.

\subsubsection{Outlier Datasets}
As an outlier dataset, earlier works have adopted 80 Million Tiny Images \cite{torralba200880}; however, it has recently been advised by \cite{prabhu2020large} that due to the presence of biases, offensive and prejudicial images it's further usage has been discontinued. Considering the ethical research practice, we therefore, use 300K Random Images, which is a de-biased subset of the same prepared by \cite{hendrycks2018deep}.

\subsubsection{OOD Datasets}
\label{sec:ood_data}
We follow the baseline works \cite{hendrycks2018deep}, \cite{liu2020energy} to adapt the common OOD dataset benchmarks. These include Textures \cite{cimpoi14describing}, LSUN-C \cite{yu15lsun}, SVHN \cite{netzer2011reading}, iSUN \cite{xu2015turkergaze}, and Places365 \cite{zhou2016places}. We only use the test sets of these data as OOD datasets.

\subsection{Baseline and S.O.T.A Approaches}
\label{sec: baseline_sota}
We compare our method with four different competitive baseline OOD detection approaches:
(1) \textbf{OE} \cite{hendrycks2018deep}, and remaining three are it's variants that follow the similar principle of model regularization by training with auxiliary outliers; 
(2) \textbf{Energy} \cite{liu2020energy} (2020) employs energy scores aligned with the probability density of inputs for OOD detection; 
(3) \textbf{MixOE} \cite{zhang2023mixture} (2023) utilizes Mixup technique to mix $\mathcal{D}_{in}$ and $\mathcal{D}_{out}^{OE}$ to further enhance model regularization; 
and (4) \textbf{DivOE} \cite{zhu2024diversified} (2024) diversifies $\mathcal{D}_{out}^{OE}$ by explicitly synthesizing more informative outliers for extrapolation during training. 
We re-implemented these methods using their publicly available source codes, following the datasets and training configurations described in Sections \ref{sec:datasets}, and \ref{sec:training_config} respectively.

\subsection{Training Configuration}
\label{sec:training_config}
In general, OE and its variants are trained in a fine-tuned scenarios. This approach is more practical because it is more common to equip deployed models with the ability to detect OOD inputs rather than training a dual task (ID classification and OOD detection) from scratch. Following a similar setup, we use pre-trained baselines for models that are available. For models that do not have a pre-trained baseline, we initially train the model from scratch using a MSP \cite{hendrycks2016baseline} objective and then utilize it for fine-tuning.

\subfour{Models and Hyperparameters: }
We train our method on four different neural network (backbone) architectures that are considered pre-eminent in image classification tasks; WideResnet \cite{zagoruyko2016wide}, Allconv \cite{salimans2016weight}, Resnet \cite{he2016deep}, and Densenet \cite{huang2017densely}. For the sake of equivalence comparison with OE \cite{hendrycks2018deep}, we use their default hyperparameters. Specifically, for WideResnet architecture we use a total of 40 layers with a widen factor of 2, and dropout rate of 0.3. Likewise, we use Allconv with 9 layers, each comprising a combination of (Conv2D - BatchNorm2D - GELU). Furthermore, we use Resnet and Densenet models with 18 and 121 layer variants respectively.
All the networks are fine-tuned on a pre-trained model upto 10 epochs using a stochastic gradient descent (SGD) optimizer with weight decay of $5e-4$, an initial learning rate of $0.001$ with cosine decay. Unlike \cite{hendrycks2018deep} that employed varying sample sizes for $\mathcal{D}_{in}$ and $\mathcal{D}_{out}^{OE}$, our approach utilizes equivalent sample sizes of $N = 128$, with a cumulative batch size of $\mathcal{B} = 2N = 256$ to enable post-hoc calculations.. The choices of $\lambda_1$, and $\lambda_2$ are both set at 0.5. Lastly, we select the value of $m \in \left\{0.1, 0.2, \hdots, 0.9\right\}$.
All experiments were conducted on multiple RTX A4000 GPU servers.

\subsection{Evaluation Metrics}
We evaluated the detection performance using several metrics: (i) \textbf{AUROC:} It measures the discriminative capability of an OOD classifier in discerning ID and OOD data. It's value ranges from 0 to 1, the latter indicating perfect distinction. (ii) \textbf{AUPR:} This metric evaluates the trade-off between precision and recall, usually under class imbalance scenarios. Higher value of AUPR indicates better detection performance. (iii) \textbf{FPR95:} This metric is significant for assessing the robustness of OOD detection under high recall conditions. Ideally, a lower value of FPR95 is desirable which indicates fewer ID samples are incorrectly classified as OOD. We also evaluate the classification performance of the ID inputs using accuracy metric represented as ID-ACC.

\section{Results Comparison}

In this section, we compare our results with the baseline and S.O.T.A methods as discussed in Section \ref{sec: baseline_sota}. Across all metrics, we report an averaged performance and a standard error value that was determined through the execution of 10 independent test trials.

\begin{table*}[ht!]
    \caption{Comparison of OOD detection results on different ID datasets fine-tuned on a WRN architecture using 300K Random Images as auxiliary outliers. Best and second best values are reported in bold, and underline respectively. Arrows represent the direction towards optimum value.}
    \centering
    \begin{subtable}{\textwidth}
    \resizebox{\textwidth}{!}{%
    \begin{tabular}{|l|c|c|c|c|c|c|c|c|}
         \hline
         
         {\backslashbox{\bf Method}{\bf ID Data}} & \multicolumn{4}{c|}{\bf CIFAR-10} & \multicolumn{4}{c|}{\bf CIFAR-100}\\
         \hline
         
          & \bf AUROC $\uparrow$ & \bf AUPR $\uparrow$& \bf FPR95 $\downarrow$ & \bf ID-ACC $\uparrow$& \bf AUROC $\uparrow$& \bf AUPR $\uparrow$ & \bf FPR95 $\downarrow$ & \bf ID-ACC $\uparrow$ \\
         
         OE \cite{hendrycks2018deep} & 98.65$\pm$0.03  & \underline{98.6$\pm$0.05} & 6.21$\pm$0.13 & 94.83$\pm$0.06 & 88.51$\pm0.15$ & 87.43$\pm0.16$ & 42.12$\pm0.44$ &  75.75$\pm$0.11\\
         Energy \cite{liu2020energy} & \underline{98.68$\pm$0.03}  & 98.49$\pm$0.05 & \underline{5.88$\pm$0.13} & 94.35$\pm$0.07 & 87.567$\pm0.06$ & 87.77$\pm0.09$ & 48.93$\pm0.19$ &  74.77$\pm$0.11\\
         MixOE \cite{zhang2023mixture} & 90.85$\pm$0.12  & 90.48$\pm$0.2 & 41.46$\pm$0.36 & 94.53$\pm$0.03 & 78.02$\pm0.22$ & 73.98$\pm0.29$ & 61.34$\pm0.38$ &  75.17$\pm$0.18\\
         DivOE \cite{zhu2024diversified} & 98.46$\pm$0.04  & 98.38$\pm$0.05 & 7.15$\pm$0.19 & \underline{95.01$\pm$0.05} & 87.42$\pm0.08$ & 86.45$\pm0.06$ & 44.21$\pm0.27$ &  \underline{75.83$\pm$0.09}\\
         MaCS & {98.79$\pm$0.02}  & {98.77$\pm$0.03} & {5.14$\pm$0.11} & {95.28$\pm$0.06} & \underline{89.43$\pm$0.08} & \underline{88.82$\pm$0.15} & \underline{41.52$\pm$0.29} & 75.53$\pm$0.07\\

         MaCS{$^*$} & \bf{98.79$\pm$0.02}  & \bf{98.77$\pm$0.03} & \bf{5.14$\pm$0.11} & \bf{95.28$\pm$0.06} & \bf{90.93$\pm$0.13} & \bf{90.28$\pm$0.21} & \bf{37.54$\pm$0.35} & \bf{76.12$\pm$0.04} \\
         \hline
         \end{tabular}}
    \caption{CIFAR-10 and CIFAR-100}
    \label{tab:table1}
    \end{subtable}
    \vspace{2em} 
    \begin{subtable}{\textwidth}
        \centering
    \resizebox{\textwidth}{!}{%
    \begin{tabular}{|l|c|c|c|c|c|c|c|c|}
         \hline
         
         {\backslashbox{\bf Method}{\bf ID Data}} & \multicolumn{4}{c|}{\bf SVHN} & \multicolumn{4}{c|}{\bf Imagenet-32}\\
         \hline
         
          & \bf AUROC $\uparrow$ & \bf AUPR $\uparrow$& \bf FPR95 $\downarrow$ & \bf ID-ACC $\uparrow$& \bf AUROC $\uparrow$& \bf AUPR $\uparrow$ & \bf FPR95 $\downarrow$ & \bf ID-ACC $\uparrow$ \\
         OE \cite{hendrycks2018deep} & 99.93$\pm$0.0  & 99.94$\pm$0.0 & 0.11$\pm$0.01 & 94.67$\pm$0.04 & 88.76$\pm0.02$ & 87.21$\pm0.02$ & 39.72$\pm0.09$ &  \underline{34.42$\pm$0.06}\\
         Energy \cite{liu2020energy} & 99.92$\pm$0.0  & 98.93$\pm$0.0 & 0.14$\pm$0.01 & 94.35$\pm$0.03 & \underline{90.88$\pm$0.02} & \underline{89.53$\pm$0.03} & \underline{31.3$\pm$0.1} &  32.26$\pm$0.06\\
         MixOE \cite{zhang2023mixture} & 96.98$\pm$0.04  & 96.44$\pm$0.08 & 13.21$\pm$0.1 & 88.59$\pm$0.36 & 72.56$\pm0.09$ & 64.0$\pm0.09$ & 59.83$\pm0.11$ &  31.66$\pm$0.05\\
         DivOE \cite{zhu2024diversified} & 99.95$\pm$0.0  & 99.95$\pm$0.0 & 0.04$\pm$0.0 & 94.66$\pm$0.02 & 90.44$\pm0.02$ & 89.14$\pm0.03$ & 35.52$\pm0.05$ &  34.34$\pm$0.05\\
         MaCS & \underline{99.97$\pm$0.0}  & \underline{99.97$\pm$0.0} & \underline{0.03$\pm$0.0} & \underline{95.2$\pm$0.03} & {91.49$\pm$0.03} & {90.47$\pm$0.03} & {30.66$\pm$0.09} &  {38.11$\pm$0.1}\\

     MaCS{$^*$} & \bf{99.98$\pm$0.0}  & \bf{99.98$\pm$0.0} & \bf{0.0$\pm$0.0} & \bf{95.4$\pm$0.02} & \bf{91.49$\pm$0.03} & \bf{90.47$\pm$0.03} & \bf{30.66$\pm$0.09} & \bf{38.11$\pm$0.1} \\
         \hline
    \end{tabular}}
    \caption{SVHN and Imagenet-32}
    \label{tab:table2}
    \end{subtable}
    \label{tab:main_1}
\end{table*}

\subsection{Detection Results}
First, we present the detection results. Here, we test on the fine-tuned methods on same backbone of Wideresnet architecture with specifications as stated in Section \ref{sec:training_config}.

\par Table \ref{tab:table1} presents the detection metrics with CIFAR-10 and CIFAR-100 as ID datasets. We present results of our method in two variants (same across all experiments, hereafter): MaCS and MaCS$^*$, the former one fine-tuned at fixed $m$ = 0.5, while the latter fine-tuned with respective optimal value of $m$ for each test setting as reported in Table. \ref{tab:table_5}. From the results in Table. \ref{tab:table1}, we can observe that MaCS$^*$ consistently outperformed all other methods across both ID datasets, not only in terms of detection metrics but also proving effective in generously classifying ID samples. MaCS was also able to obtain good detection performance coming second to MaCS$^*$ with CIFAR-100 ID data. The key reason for the improved performance can be attributed to the weighted penalization feature of MaCS. Because the model is trained to focus entirely on the score explosions, it becomes apparent that the model learns to restrict OOD scores to be smaller than that of ID scores. The comprehensive results on CIFAR ID benchmarks for each test OOD dataset evaluated under different methods with different backbone architectures are listed in Table \ref{tab:table_3}. 

\par Similarly, in Table \ref{tab:table2}, we compare our results by changing the ID inputs from CIFAR datasets to SVHN and Imagenet-32 but fine-tuned on the same architecture. Analyzing the results, it is evident that MaCS$^*$ performance remains superior regardless of change in $\mathcal{D}_{in}$. For SVHN, MaCS$^*$ reports FPR95 value to be as low as zero, while for large-scale Imagenet-32 we beat OE, and Energy \cite{liu2020energy}, the second best method by 4.58 \%, and 0.64\% respectively. Note that for tests conducted with SVHN and Imagenet-32 as ID datasets, the results represent average scores across all OOD datasets except SVHN. Considering only the confidence score based supplementary constraint to conventional OE's objective, the gain in OOD detection performance is substantial. Interestingly, it is noteworthy to realize that MaCS and MaCS$^*$ were able to outperform relatively sophisticated methods such as MixOE \cite{zhang2023mixture} and DivOE \cite{zhu2024diversified}. This demonstrates the effectiveness of the proposed method which in addition to being conceptually simpler also yields exquisite performance. The comprehensive results on SVHN and Imagenet-32 ID benchmarks for each test OOD dataset evaluated under different methods with different backbone architecture are listed in Table \ref{tab:table_4}. 

\par On other note, while training to distinguish ID and OOD samples based on their confidence scores, our method simultaneously learns to make the inter-class decision boundary of ID samples more compact, leading to fewer classification errors. The rationale behind this is that, with the cost function being penalized for every score explosion, the model takes wise decision in mapping inputs to corresponding distributions while keeping the loss value down throughout.

\begin{table*}[ht!]
    \caption{Comprehensive OOD detection results comparison of MaCS on different ID datasets with S.O.T.A methods. All methods are trained on a WRN architecture. Best, and second best results are represented in bold and underline respectively.}
    \centering
    \begin{subtable}{\textwidth}
  \resizebox{\textwidth}{!}{%
  \begin{tabular}{|l|c|ccc|ccc|ccc|ccc|}
    \hline
    & &\multicolumn{6}{c|}{\bf CIFAR-10} & \multicolumn{6}{c|}{\bf CIFAR-100}\\
    \cline{3-14}
    & & \multicolumn{3}{c|}{\bf WRN} & \multicolumn{3}{c|}{\bf Allconv} & \multicolumn{3}{c|}{\bf WRN} & \multicolumn{3}{c|}{\bf Allconv} \\
    
    \bf OOD Data & \textbf{Methods} & \textbf{AUROC} $\uparrow$ & \textbf{AUPR} $\uparrow$ & \textbf{FPR95} $\downarrow$ & \textbf{AUROC} $\uparrow$ & \textbf{AUPR} $\uparrow$ & \textbf{FPR95} $\downarrow$ & \textbf{AUROC} $\uparrow$ & \textbf{AUPR} $\uparrow$ & \textbf{FPR95} $\downarrow$ & \textbf{AUROC} $\uparrow$ & \textbf{AUPR} $\uparrow$ & \textbf{FPR95} $\downarrow$ \\
    \hline

    \multirow{4}{2em}{Textures} & OE \cite{hendrycks2018deep} &  98.40±0.05 & 98.35±0.06 & 8.49±0.28 & 97.67±0.04 & 97.73±0.04 & 13.21±0.30& 86.22±0.18 & 85.53±0.21 & 49.84±0.64 & 80.82±0.14 & 79.22±0.22 & 60.12±0.33 \\ 
    
    & Energy \cite{liu2020energy}  & \underline{98.68±0.03} & \underline{98.56±0.04} & \underline{6.41±0.28} & 97.27±0.04 & 97.44±0.04 & 14.32±0.31 & 84.39±0.07 & 85.05±0.08 & 59.49±0.45 & 77.67±0.19 & 77.59±0.20& 68.19±0.85 \\
    & MixOE \cite{zhang2023mixture}  & 87.91±0.12 & 87.87±0.20& 60.21±0.70& 93.04±0.07 & 92.19±0.07 & 26.58±0.32 & 76.15±0.24 & 72.33±0.32 & 68.73±0.50& 74.95±0.16 & 71.17±0.19 & 71.33±0.61 \\
    
    & DivOE \cite{zhu2024diversified}  & 98.25±0.05 & 98.23±0.05 & 9.37±0.31 & 97.63±0.05 & 97.74±0.06 & 13.78±0.34 & {87.08±0.11} & {86.97±0.1} & {48.37±0.36} & {81.34±0.19} & \underline{81.02±0.2} & {60.05±0.48}\\
    & MaCS  & 98.74±0.03 & 98.76±0.03 & 5.07±0.16 & \underline{98.29±0.04} & \underline{98.28±0.04} & \underline{9.60±0.23} & \underline{88.15±0.11} & \underline{88.14±0.12} & \underline{46.89±0.43} & \underline{81.84±0.16} & {80.42±0.25} & \underline{58.73±0.57}\\

    & MaCS$^*$ & \bf{98.74±0.03} & \bf{98.76±0.03} & \bf{5.07±0.16} & \bf 98.69±0.02 & \bf 98.72±0.03 & \bf 7.68±0.18 & \bf{88.61±0.18} & \bf 88.49±0.23 & \bf 46.00±0.60& \bf83.81±0.20& \bf 83.31±0.14 & \bf 57.18±0.69\\
    
    \hline
    \multirow{4}{2em}{iSUN} & OE \cite{hendrycks2018deep} &  \underline{99.05±0.04} & \underline{98.88±0.07} & 4.60±0.16 & 98.17±0.04 & 98.03±0.04 & 9.30±0.25 & 84.79±0.14 & 83.38±0.14 & 52.81±0.49 & \underline{68.81±0.2} & \underline{70.01±0.22} & 82.22±0.39\\ 
    
    & Energy \cite{liu2020energy}  & 99.10±0.03 & 98.83±0.05 & \underline{3.38±0.14} & 96.50±0.06 & 96.24±0.07 & 15.54±0.32 & \underline{86.95±0.15} & \underline{86.99±0.19} & \underline{51.36±0.45} & 63.90±0.18 & 63.64±0.26 & \underline{78.28±0.32}\\
    
    & MixOE \cite{zhang2023mixture}  & 89.78±0.15 & 89.15±0.22 & 43.64±1.04 & 96.63±0.04 & 96.15±0.06 & 14.07±0.25 & 70.31±0.32 & 65.02±0.37 & 74.29±0.40& 66.31±0.16 & 65.63±0.21 & 85.06±0.26\\
    
    & DivOE \cite{zhu2024diversified}  & 98.95±0.03 & 98.78±0.04 & 5.19±0.21 & \underline{98.47±0.04} & \underline{98.43±0.05} & {8.60±0.22} & 81.40±0.16 & 79.67±0.17 & 57.54±0.71 & 66.47±0.11 & 68.14±0.12 & 83.32±0.34\\
    
    & MaCS  & 99.24±0.02 & 99.11±0.04 & 3.28±0.11 & {98.46±0.04} & {98.24±0.06} & \underline{7.62±0.21} & {86.58±0.13} & {85.73±0.22} & \underline{49.90±0.3} & {67.73±0.24} & {68.42±0.26} & {81.24±0.23}\\

    & MaCS$^*$ & \bf 99.24±0.02 & \bf 99.11±0.04 & \bf 3.28±0.11 & \bf 98.62±0.04 & \bf 98.46±0.05 & \bf 6.86±0.18 & \bf 89.75±0.14 & \bf 88.39±0.25 & \bf 39.75±0.58 & \bf 74.33±0.21 & \bf 74.38±0.21 & \bf 72.92±0.47\\
    
    \hline
    \multirow{4}{2em}{LSUN-C} & OE \cite{hendrycks2018deep} &  \bf99.74±0.01 & \bf99.74±0.01 & \bf1.10±0.07 & {99.65±0.01} & {99.65±0.01} & {1.66±0.09} & \bf97.00±0.08 & \bf96.94±0.07 & \bf14.94±0.61 & {96.22±0.06} & {96.30±0.08} & {20.49±0.34}\\ 
    
    & Energy \cite{liu2020energy}  & 99.55±0.02 & 99.34±0.04 & \underline{1.46±0.09} & 99.43±0.02 & 99.41±0.03 & 3.20±0.17 & 94.67±0.08 & 95.06±0.09 & 31.47±0.27 & 93.80±0.09 & 94.36±0.08 & 35.07±0.72\\
    
    & MixOE \cite{zhang2023mixture}  & 97.30±0.09 & 97.07±0.12 & 11.92±0.25 & 97.99±0.03 & 97.65±0.05 & 9.36±0.15 & 92.08±0.14 & 91.27±0.16 & 31.39±0.56 & 91.67±0.05 & 90.61±0.09 & 30.07±0.29\\
    
    & DivOE \cite{zhu2024diversified}  & 99.64±0.02 & \underline{99.64±0.02} & 1.80±0.14 & 99.56±0.02 & 99.56±0.02 & 2.38±0.16 & \underline{96.54±0.05} & \underline{96.51±0.05} & \underline{17.35±0.36} & 95.35±0.08 & 95.61±0.07 & 25.86±0.58\\
    
    & MaCS  & {99.64±0.01} & 99.63±0.02 & 1.62±0.08 & \underline{99.73±0.01} & \underline{99.72±0.01} & \bf{1.11±0.06} & 95.84±0.11 & 95.59±0.14 & 19.70±0.64 & \underline{96.36±0.09} & \underline{96.36±0.11} & \underline{18.37±0.47}\\

    & MaCS$^*$ & \underline{99.64±0.01} & 99.63±0.02 & 1.62±0.08 & \bf 99.75±0.01 & \bf 99.75±0.02 & \underline{1.16±0.09} & 96.03±0.10& 95.72±0.14 & 18.39±0.71 & \bf 96.74±0.05 & \bf 96.82±0.03 & \bf 17.02±0.82\\
    
    \hline
    \multirow{4}{2em}{SVHN} & OE \cite{hendrycks2018deep} & \bf99.35±0.03 & \underline{99.19±0.06} & \bf2.14±0.08 & {98.99±0.03} & {98.92±0.04} & {4.98±0.23} & 88.14±0.20& 85.73±0.23 & {42.01±0.39} & {85.78±0.17} & {81.06±0.29} & {41.88±0.34}\\ 
    
    & Energy \cite{liu2020energy}  &  99.00±0.04 & 98.42±0.07 & \underline{2.62±0.09} & 96.36±0.07 & 95.43±0.11 & 11.94±0.26 & {89.39±0.09} & \underline{89.14±0.12} & 43.40±0.37 & 80.65±0.14 & 76.10±0.25 & 49.97±0.41\\
    
    & MixOE \cite{zhang2023mixture}  &  91.68±0.19 & 90.71±0.27 & 31.18±0.97 & 89.52±0.09 & 84.02±0.18 & 28.69±0.30& 74.84±0.28 & 67.66±0.34 & 63.05±0.44 & 76.51±0.23 & 68.05±0.30& 53.54±0.62\\
    
    & DivOE \cite{zhu2024diversified}  &  99.11±0.03 & 98.85±0.06 & 3.23±0.14 & 98.00±0.05 & 97.71±0.07 & 9.27±0.25 & 86.89±0.14 & 84.68±0.13 & 44.44±0.42 & 83.47±0.22 & 78.62±0.32 & 45.34±0.3\\
    
    & MaCS  &  {99.31±0.02} & {99.14±0.04} & 2.72±0.14 & \underline{99.49±0.02} & \underline{99.44±0.02} & \underline{2.26±0.05} & \underline{90.01±0.1} & {88.97±0.16} & \underline{40.09±0.43} & \underline{88.11±0.12} & \underline{84.47±0.29} & \underline{39.61±0.4}\\

    & MaCS$^*$ & \underline{99.31±0.02} & \bf{99.14±0.04} & 2.72±0.14 & \bf 99.65±0.02 & \bf 99.64±0.02 & \bf 1.63±0.11 & \bf 93.03±0.13 & \bf 92.40±0.20& \bf 32.68±0.50& \bf 88.93±0.11 & \bf 86.11±0.13 & \bf 39.76±0.74\\
    
    \hline
    \multirow{4}{2em}{Places365} & OE \cite{hendrycks2018deep} & 96.73±0.07 & 96.86±0.08 & \underline{14.74±0.4} & {95.03±0.08} & {95.00±0.09} & {21.76±0.54} & {86.42±0.25} & {85.56±0.26} & \underline{50.97±0.9} & {83.72±0.16} & {82.56±0.23} &{55.82±0.45}\\ 
    & Energy \cite{liu2020energy} & \bf97.06±0.08 & \bf97.29±0.07 & 15.50±0.52 & 94.39±0.06 & 94.46±0.07 & 24.72±0.32 & 82.93±0.08 & 82.61±0.10& 58.95±0.49 & 79.80±0.18 & 78.96±0.20& 61.28±0.52\\
    & MixOE \cite{zhang2023mixture} & 87.56±0.20& 87.63±0.27 & 60.35±0.90& 90.41±0.11 & 88.86±0.15 & 35.43±0.46 & 76.75±0.30& 73.61±0.41 & 69.26±0.69 & 79.99±0.18 & 76.46±0.26 & 60.52±0.57\\
    & DivOE \cite{zhu2024diversified} & 96.34±0.11 & 96.41±0.10& 16.16±0.54 & 94.82±0.04 & 94.80±0.05 & 22.20±0.37 & 85.18±0.10& 84.44±0.07 & 53.36±0.42 & 83.16±0.17 & 82.13±0.17 & 56.95±0.52\\
    & MaCS & {97.03±0.08} & {97.24±0.07} & 13.00±0.45 & \underline{95.98±0.07} & \underline{96.08±0.08} & \underline{18.96±0.34} & \underline{86.56±0.14} & \underline{85.67±0.18} & {51.04±0.74} & \underline{84.26±0.2} & \underline{82.96±0.21} & \underline{54.02±0.59}\\

    & MaCS$^*$ & \underline{97.03±0.08} & \underline{97.24±0.07} & \bf{13.00±0.45} & \bf 96.43±0.12 & \bf 96.59±0.11 & \bf 16.89±0.61 & \bf 87.23±0.20& \bf 86.41±0.29 & \bf 50.86±0.75 & \bf 84.88±0.20& \bf 83.87±0.19 & \bf 53.65±0.58\\
      
  \hline
  \end{tabular}}
    \caption{CIFAR-10 and CIFAR-100}
    \label{tab:table_3}
    \end{subtable}
    \vspace{1em} 
    \begin{subtable}{\textwidth}
        \centering
  \resizebox{\textwidth}{!}{%
  \begin{tabular}{|l|c|ccc|ccc|ccc|ccc|}
    \hline
    & &\multicolumn{6}{c|}{\bf SVHN} & \multicolumn{6}{c|}{\bf Imagenet32}\\
    \cline{3-14}
    & & \multicolumn{3}{c|}{\bf WRN} & \multicolumn{3}{c|}{\bf Allconv} & \multicolumn{3}{c|}{\bf WRN} & \multicolumn{3}{c|}{\bf Allconv} \\
    
    \bf OOD Data & \textbf{Methods} & \textbf{AUROC} $\uparrow$ & \textbf{AUPR} $\uparrow$ & \textbf{FPR95} $\downarrow$ & \textbf{AUROC} $\uparrow$ & \textbf{AUPR} $\uparrow$ & \textbf{FPR95} $\downarrow$ & \textbf{AUROC} $\uparrow$ & \textbf{AUPR} $\uparrow$ & \textbf{FPR95} $\downarrow$ & \textbf{AUROC} $\uparrow$ & \textbf{AUPR} $\uparrow$ & \textbf{FPR95} $\downarrow$ \\
    \hline

    \multirow{4}{2em}{Textures} & OE \cite{hendrycks2018deep} &  99.75±0.00& 99.79±0.00& 0.43±0.03 & 99.96±0.00& 99.96±0.00& 0.02±0.01 & 80.16±0.03 & 7.24±0.03 & 75.17±0.09 & 84.27±0.04 & 82.08±0.04 & 71.51±0.12 \\ 
    
    & Energy \cite{liu2020energy} &  99.71±0.00& 99.74±0.00&0.54±0.03 & 99.97±0.00& 99.97±0.00& \underline{0.01±0.0} & \underline{86.78±0.02} & \underline{85.81±0.03} & 67.39±0.10& 76.57±0.05 & 74.65±0.04 & 82.13±0.08 \\ 
    
   & MixOE \cite{zhang2023mixture} &  94.98±0.05 & 94.39±0.10&22.32±0.10& 93.75±0.05 & 92.32±0.08 & 22.01±0.20& 55.90±0.11 & 37.25±0.08 & 85.75±0.10& 59.11±0.07 & 42.78±0.07 & 85.43±0.11 \\ 
    
    & DivOE \cite{zhu2024diversified} &  99.85±0.00& 99.86±0.01 &0.10±0.01 & \underline{99.98±0.0} & \underline{99.98±0.0} & {0.01±0.01} & 86.59±0.03 & 85.18±0.04 & \underline{67.12±0.11} & \underline{85.41±0.03} & \underline{84.01±0.04} & \underline{70.45±0.08} \\ 
    
    & MaCS &  \underline{99.87±0.0} & \underline{99.88±0.0} &0.11±0.01 & 99.95±0.00& 99.96±0.00& 0.00±0.00& 87.43±0.04 & 86.19±0.04 & 64.63±0.20& 83.56±0.03 & 81.44±0.04 & 73.58±0.13\\ 

    & MaCS$^*$ &  \bf 99.91±0.00& \bf 99.92±0.00&\bf 0.00±0.00& \bf 99.99±0.00& \bf 99.99±0.00& \bf 0.00±0.00& \bf 87.43±0.04 & \bf 86.19±0.04 & \bf 64.63±0.20& \bf 88.73±0.03 & \bf 87.46±0.06 & \bf 63.80±0.17\\

    \hline
    \multirow{4}{2em}{iSUN} & OE \cite{hendrycks2018deep} &  100.00±0.00& 100.00±0. &0.00±0.00& 100.00±0.00& 100.00±0.00& 0.00±0.00& \underline{70.98±0.06} & {64.30±0.06} & \bf{73.99±0.09} & {57.70±0.1} & 51.63±0.08 & 85.69±0.11 \\ 
    
    & Energy \cite{liu2020energy} &  100.00±0.00& 100.00±0.00&0.00±0.00& 100.00±0.00& 100.00±0.00& 0.00±0.00& 70.47±0.06 & 63.90±0.10& 73.16±0.13 & \underline{62.10±0.09} & \underline{57.62±0.07} & \underline{84.87±0.08} \\ 
    
   & MixOE \cite{zhang2023mixture} &  \underline{98.31±0.04} & \underline{97.86±0.07} &\underline{6.89±0.13} & \underline{97.15±0.03} & \underline{96.31±0.05} & \underline{10.85±0.1} & 64.99±0.11 & 54.92±0.10& 71.72±0.16 & 57.13±0.08 & 49.72±0.06 & 86.42±0.11 \\ 
    
    & DivOE \cite{zhu2024diversified} &  100.00±0.00& 99.99±0.00&0.01±0.00& 100.00±0.00& 100.00±0.00& 0.00±0.00& 70.86±0.07 & \underline{64.39±0.11} & 73.48±0.13 & 57.49±0.07 & 51.88±0.07 & 86.12±0.09 \\ 
    & MaCS &  100.00±0.00& 100.00±0.00&0.00±0.00& 100.00±0.00& 100.00±0.00& 0.00±0.00& 72.65±0.09 & 68.07±0.12 & 73.99±0.15 & 56.95±0.08 & 51.74±0.07 & 85.42±0.10\\ 
    & MaCS$^*$ &  \bf 100.00±0.00& \bf 100.00±0.00& \bf 0.00±0.00& \bf 100.00±0.00& \bf 100.00±0.00& \bf 0.00±0.00& \bf 72.65±0.09 & \bf 68.07±0.12 & \underline{73.99±0.15} & \bf 64.22±0.07 & \bf 58.91±0.12 & \bf 82.96±0.13\\

    \hline
    \multirow{4}{2em}{LSUN-C} & OE \cite{hendrycks2018deep} &  \underline{99.98±0.0} & \underline{99.98±0.0} &\underline{0.01±0.0} & 100.00±0.00& 100.00±0.00& 0.00±0.00& 99.66±0.00& 99.74±0.00& \underline{0.04±0.01} & 99.69±0.00& 99.75±0.00& 0.14±0.01 \\ 
    
    & Energy \cite{liu2020energy} &  100.00±0.00& 100.00±0.00&0.00±0.00& 100.00±0.00& 100.00±0.00& 0.00±0.00& \underline{99.72±0.0} & \underline{99.78±0.0} & 0.05±0.00& 99.61±0.00& 99.70±0.00& 0.06±0.01 \\ 
    
   & MixOE \cite{zhang2023mixture} &  98.31±0.04 & 97.86±0.07 &6.89±0.13 & 96.38±0.03 & 95.72±0.04 & \underline{14.25±0.14} & 88.62±0.06 & 86.01±0.10& 38.45±0.12 & 93.30±0.03 & 93.07±0.04 & 30.82±0.08 \\ 
    
    & DivOE \cite{zhu2024diversified} &  99.97±0.00& 99.97±0.01 &0.02±0.01 & 100.00±0.00& 100.00±0.00& 0.00±0.00& 99.50±0.00& 99.61±0.00& 0.16±0.01 & 99.65±0.00& 99.73±0.00& 0.20±0.01 \\ 
    & MaCS &  100.00±0.00& 100.00±0.00&0.00±0.00& \underline{99.99±0.0} & \underline{99.99±0.0} & {0.00±0.0} & 99.81±0.00& 99.86±0.00& 0.04±0.00& \underline{99.75±0.0} & \underline{99.81±0.0} & \underline{0.06±0.01} \\ 
    & MaCS$^*$ &  \bf 100.00±0.00& \bf 100.00±0.00&\bf 0.00±0.00& \bf 100.00±0.00& \bf 100.00±0.00& \bf 0.00±0.00& \bf 99.81±0.00& \bf 99.86±0.00& \bf 0.04±0.00& \bf 99.93±0.00& \bf 99.94±0.00& \bf 0.04±0.01\\

    \hline
    \multirow{4}{2em}{Places365} & OE \cite{hendrycks2018deep} &  99.99±0.00& 9.99±0.00&\underline{0.01±0.0} & 100.00±0.00& 100.00±0.00& 0.00±0.00& 93.00±0.06 & 94.77±0.04 & 49.38±0.37 & 96.47±0.03 & 97.18±0.02 & 23.34±0.28 \\ 
    
    & Energy \cite{liu2020energy} &  99.98±0.00& 99.98±0.00&0.00±0.00& 100.00±0.00& 100.00±0.00& 0.00±0.00& \underline{97.43±0.03} & \underline{98.15±0.03} & \underline{15.88±0.45} & 97.01±0.03 & 97.78±0.02 & 20.76±0.34 \\ 
    
   & MixOE \cite{zhang2023mixture} &  97.24±0.05 & 96.44±0.09 &10.87±0.16 & \underline{96.37±0.05} & \underline{95.21±0.06} & \underline{13.02±0.18} & 61.27±0.12 & 58.17±0.12 & 85.17±0.14 & 59.95±0.10& 56.62±0.08 & 85.34±0.18 \\ 
    
    & DivOE \cite{zhu2024diversified} &  99.99±0.00& 99.99±0.00&0.02±0.01 & 100.00±0.00& 100.00±0.00& 0.00±0.00& 95.26±0.02 & 96.50±0.02 & 36.82±0.19 & 97.32±0.02 & 97.88±0.02 & 16.51±0.12 \\ 
    & MaCS &  \underline{99.99±0.0} & \underline{99.99±0.0} &0.00±0.00& 100.00±0.00& 100.00±0.00& 0.00±0.00& 97.57±0.02 & 98.21±0.02 & 14.65±0.31 & \underline{97.75±0.02} & \underline{98.24±0.02} & \underline{13.34±0.21} \\ 
    & MaCS$^*$ &  \bf 100.00±0.00& \bf 100.00±0.00&\bf 0.00±0.00& \bf 100.00±0.00& \bf 100.00±0.00& \bf 0.00±0.00& \bf 97.57±0.02 & \bf 98.21±0.02 & \bf 14.65±0.31 & \bf 99.36±0.02 & \bf 99.46±0.02 & \bf 0.92±0.05\\
    
  \hline
  \end{tabular}}
    \caption{SVHN and Imagenet-32}
    \label{tab:table_4}
    \end{subtable}
    
    \label{tab:main_2}
\end{table*}

\subsection{Confidence Scores Disparity between ID and OOD Data}
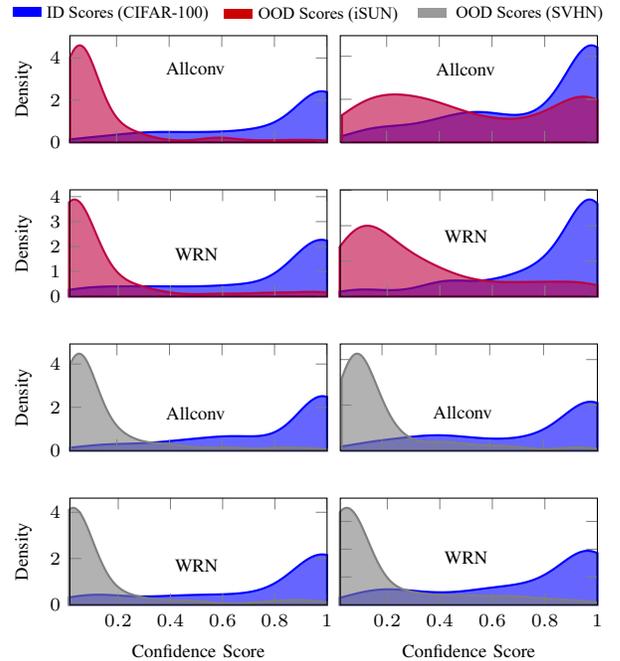
\begin{figure}[ht!]
\centering
\includegraphics[width = 0.5\textwidth]{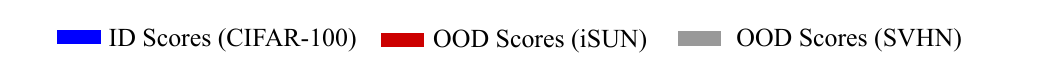}
\begin{tikzpicture}

\begin{groupplot}[
    group style={
        group size=2 by 4,
        xlabels at=edge bottom,
        ylabels at=edge left,
        xticklabels at=edge bottom,
        yticklabels at=edge left,
        vertical sep=18pt,
        horizontal sep=5pt,
    },
    width=5cm,
    height=3cm,
    xlabel={Confidence Score},
    ylabel={Density},
    grid=none, 
    axis line style={draw=black},
    xmax=1, 
    ymin=0, 
    axis on top,
    clip=false,
    xtick align=inside,
    ytick align=inside,
    tick label style={font=\scriptsize},
    label style={font=\scriptsize},
    title style = {font=\scriptsize},
    legend style={at={(.5,-0.1)},anchor=south}
    legend to name=grouplegend,
    axis line style={draw=none}, 
    x axis line style={draw=black}, 
    y axis line style={draw=black} 
]

\nextgroupplot[xmin=0.023]
\addplot[blue, thick, fill=blue, fill opacity=0.6] 
    table [x=x, y=id_density, col sep=space] {data/isun_macs_allconv.dat} \closedcycle;

\addplot[purple, thick, fill=purple, fill opacity=0.6] 
    table [x=x, y=ood_density, col sep=space] {data/isun_macs_allconv.dat} \closedcycle;
\node[font=\scriptsize] at (axis cs: 0.5,3.5) {Allconv};

\nextgroupplot[xmin=0.024]
\addplot[blue, thick, fill=blue, fill opacity=0.6] 
    table [x=x, y=id_density, col sep=space] {data/isun_oe_allconv.dat} \closedcycle;

\addplot[purple, thick, fill=purple, fill opacity=0.6] 
    table [x=x, y=ood_density, col sep=space] {data/isun_oe_allconv.dat} \closedcycle;
\node[font=\scriptsize] at (axis cs: 0.5,1.7) {Allconv};

\nextgroupplot[xmin=0.015]
\addplot[blue, thick, fill=blue, fill opacity=0.6] 
    table [x=x, y=id_density, col sep=space] {data/isun_macs_wrn.dat} \closedcycle;

\addplot[purple, thick, fill=purple, fill opacity=0.6] 
    table [x=x, y=ood_density, col sep=space] {data/isun_macs_wrn.dat} \closedcycle;
\node[font=\scriptsize] at (axis cs: 0.5,1.7) {WRN};

\nextgroupplot[xmin=0.021]
\addplot[blue, thick, fill=blue, fill opacity=0.6] 
    table [x=x, y=id_density, col sep=space] {data/isun_oe_wrn.dat} \closedcycle;

\addplot[purple, thick, fill=purple, fill opacity=0.6] 
    table [x=x, y=ood_density, col sep=space] {data/isun_oe_wrn.dat} \closedcycle;
\node[font=\scriptsize] at (axis cs: 0.5,1.7) {WRN};
    
\nextgroupplot[xmin=0.021]
\addplot[blue, thick, fill=blue, fill opacity=0.6] 
    table [x=x, y=id_density, col sep=space] {data/svhn_macs_allconv.dat} \closedcycle;

\addplot[gray, thick, fill=gray, fill opacity=0.6] 
    table [x=x, y=ood_density, col sep=space] {data/svhn_macs_allconv.dat} \closedcycle;
\node[font=\scriptsize] at (axis cs: 0.5,1.7) {Allconv};

\nextgroupplot[xmin=0.045]
\addplot[blue, thick, fill=blue, fill opacity=0.6] 
    table [x=x, y=id_density, col sep=space] {data/svhn_oe_allconv.dat} \closedcycle;

\addplot[gray, thick, fill=gray, fill opacity=0.6] 
    table [x=x, y=ood_density, col sep=space] {data/svhn_oe_allconv.dat} \closedcycle;
\node[font=\scriptsize] at (axis cs: 0.5,1.7) {Allconv};

\nextgroupplot[xmin=0.015]
\addplot[blue, thick, fill=blue, fill opacity=0.6] 
    table [x=x, y=id_density, col sep=space] {data/svhn_macs_wrn.dat} \closedcycle;

\addplot[gray, thick, fill=gray, fill opacity=0.6] 
    table [x=x, y=ood_density, col sep=space] {data/svhn_macs_wrn.dat} \closedcycle;
\node[font=\scriptsize] at (axis cs: 0.5,1.7) {WRN};

\nextgroupplot[xmin=0.021, legend to name=kde]
\addplot[blue, thick, fill=blue, fill opacity=0.6] 
    table [x=x, y=id_density, col sep=space] {data/svhn_oe_wrn.dat} \closedcycle;

\addplot[gray, thick, fill=gray, fill opacity=0.6] 
    table [x=x, y=ood_density, col sep=space] {data/svhn_oe_wrn.dat} \closedcycle;
\node[font=\scriptsize] at (axis cs: 0.5,1.7) {WRN};
\end{groupplot}

\end{tikzpicture}
\caption{KDE plot of confidence scores for two OOD test data: iSUN and SVHN against CIFAR-100 ID data trained on a WRN architecture. Left column plots are for MaCS, and right column plots are for OE.}
\label{fig:figure_3}
\end{figure}

MaCS's objective is to penalize score explosions, with the aim of increasing the disparity between ID and OOD scores. This is intended to make the separation between the two more apparent when thresholding with \eqref{eq:2}. To illustrate this property of MaCS, we trained two different backbone architectures, WRN and Allconv, using CIFAR-100 as ID data. We then plotted the Kernel Density Estimation (KDE) plot of the confidence scores for two OOD test datasets: SVHN and iSUN, as shown in Fig. \ref{fig:figure_3}. As can be seen from the figure, the confidence scores for ID data are higher and close to 1, while those for OOD data are close to 0. Interestingly, we can also see that the overlap between these scores for MaCS is lower than that of OE, indicating that MaCS is better at distinguishing between ID and OOD samples. Given that MaCS penalizes score explosions and limits them to a defined margin, (i) OOD scores tend to be lower than their ID counterpart, and (ii) a sufficient gap (equivalent to $m$) between ID and OOD scores is ensured.

\section{Ablation Study}
\label{sec:ablation}
In this section we describe multiple experiments performed to evaluate the contributions made by the individual components of the proposed method.

\subsection{Effect of Margin on the Detection Performance}
\label{sec:margin}
\begin{figure}[ht!]
\begin{center}
\begin{tikzpicture}
\begin{groupplot}[
group style={
    group name=my plots,
    group size=3 by 4,
    xlabels at=edge bottom,
    ylabels at=edge left,
    horizontal sep=1.08cm,
    vertical sep=1.2cm},
legend style={at={(.5,0.9)},anchor=north east, legend columns = 4}, grid=both, grid style=dashed, tick label style={font=\tiny}, label style={font=\tiny},legend style={font=\scriptsize}, 
width=0.39\linewidth, height = 0.35\linewidth,
]
\nextgroupplot[xtick={0, 0.2, 0.4, 0.6, 0.8, 1.0}, xticklabel style={rotate=45, anchor=east}, ylabel=AUROC, xlabel=Margin]

\addplot[mark=x,blue]
    coordinates{
    (0.0, 97.93)
     (0.1, 98.1)
     (0.2, 98.17)
     (0.3, 98.16)
     (0.4, 98.26)
     (0.5, 98.39)
     (0.6, 98.5)
     (0.7, 98.45)
     (0.8, 98.53)
     (0.9, 98.63)};
\addplot[mark=x,black]
    coordinates {
    (0.0, 98.54)
     (0.1, 98.73)
     (0.2, 98.79)
     (0.3, 98.7)
     (0.4, 98.81)
     (0.5, 98.79)
     (0.6, 98.67)
     (0.7, 98.46)
     (0.8, 98.56)
     (0.9, 98.43)};

\nextgroupplot[xtick={0, 0.2, 0.4, 0.6, 0.8, 1.0}, xticklabel style={rotate=45, anchor=east}, ylabel=AUPR, xlabel=Margin]

\addplot[mark=x,blue]
    coordinates{
(0.0, 97.9)
 (0.1, 98.04)
 (0.2, 98.13)
 (0.3, 98.14)
 (0.4, 98.21)
 (0.5, 98.35)
 (0.6, 98.48)
 (0.7, 98.41)
 (0.8, 98.53)
 (0.9, 98.63)};
\addplot[mark=x,black]
    coordinates{
(0.0, 98.43)
 (0.1, 98.65)
 (0.2, 98.78)
 (0.3, 98.65)
 (0.4, 98.79)
 (0.5, 98.77)
 (0.6, 98.53)
 (0.7, 98.28)
 (0.8, 98.21)
 (0.9, 98.15)};


\nextgroupplot[xtick={0, 0.2, 0.4, 0.6, 0.8, 1.0},xticklabel style={rotate=45, anchor=east}, ylabel=FPR95, xlabel=Margin]

\addplot[mark=x,blue]
    coordinates{
(0.0, 10.1)
 (0.1, 9.28)
 (0.2, 9.01)
 (0.3, 9.1)
 (0.4, 8.5)
 (0.5, 7.91)
 (0.6, 7.4)
 (0.7, 7.67)
 (0.8, 7.45)
 (0.9, 6.85)};
\addplot[mark=x,black]
    coordinates{
(0.0, 6.8)
 (0.1, 5.84)
 (0.2, 5.62)
 (0.3, 5.98)
 (0.4, 5.44)
 (0.5, 5.14)
 (0.6, 5.68)
 (0.7, 6.64)
 (0.8, 5.34)
 (0.9, 5.88)};

\nextgroupplot[xtick={0, 0.2, 0.4, 0.6, 0.8, 1.0}, xticklabel style={rotate=45, anchor=east}, ylabel=AUROC, xlabel=Margin]

\addplot[mark=x,blue]
    coordinates{
(0.0, 82.95)
 (0.1, 84.1)
 (0.2, 83.96)
 (0.3, 84.4)
 (0.4, 84.47)
 (0.5, 83.66)
 (0.6, 84.88)
 (0.7, 84.29)
 (0.8, 85.08)
 (0.9, 85.74)};
\addplot[mark=x,black]
    coordinates {
(0.0, 88.37)
 (0.1, 88.71)
 (0.2, 89.34)
 (0.3, 89.79)
 (0.4, 89.45)
 (0.5, 89.43)
 (0.6, 90.51)
 (0.7, 89.88)
 (0.8, 90.93)
 (0.9, 89.64)};

\nextgroupplot[xtick={0, 0.2, 0.4, 0.6, 0.8, 1.0}, xticklabel style={rotate=45, anchor=east}, ylabel=AUPR, xlabel=Margin]

\addplot[mark=x,blue]
    coordinates{
(0.0, 81.57)
 (0.1, 82.85)
 (0.2, 83.24)
 (0.3, 83.51)
 (0.4, 83.61)
 (0.5, 82.53)
 (0.6, 84.14)
 (0.7, 83.04)
 (0.8, 84.1)
 (0.9, 84.9)};
\addplot[mark=x,black]
    coordinates{
(0.0, 87.29)
 (0.1, 87.73)
 (0.2, 88.59)
 (0.3, 89.2)
 (0.4, 88.78)
 (0.5, 88.82)
 (0.6, 90.04)
 (0.7, 89.35)
 (0.8, 90.28)
 (0.9, 89.11)};


\nextgroupplot[xtick={0, 0.2, 0.4, 0.6, 0.8, 1.0}, xticklabel style={rotate=45, anchor=east}, ylabel=FPR95, xlabel=Margin]

\addplot[mark=x,blue]
    coordinates{
(0.0, 51.67)
 (0.1, 50.4)
 (0.2, 49.98)
 (0.3, 49.33)
 (0.4, 49.45)
 (0.5, 50.39)
 (0.6, 49.55)
 (0.7, 50.53)
 (0.8, 48.35)
 (0.9, 48.11)};
\addplot[mark=x,black]
    coordinates{
(0.0, 42.34)
 (0.1, 41.89)
 (0.2, 40.29)
 (0.3, 39.41)
 (0.4, 40.81)
 (0.5, 41.52)
 (0.6, 38.62)
 (0.7, 40.87)
 (0.8, 37.54)
 (0.9, 42.12)};

\nextgroupplot[xtick={0, 0.2, 0.4, 0.6, 0.8, 1.0}, xticklabel style={rotate=45, anchor=east}, ylabel=AUROC, xlabel=Margin]

\addplot[mark=x,blue]
    coordinates{
(0.0, 99.99)
 (0.1, 99.97)
 (0.2, 99.99)
 (0.3, 99.99)
 (0.4, 99.99)
 (0.5, 99.99)
 (0.6, 99.99)
 (0.7, 99.99)
 (0.8, 100.0)
 (0.9, 99.99)};
\addplot[mark=x,black]
    coordinates {
(0.0, 99.92)
 (0.1, 99.93)
 (0.2, 99.96)
 (0.3, 99.96)
 (0.4, 99.97)
 (0.5, 99.97)
 (0.6, 99.97)
 (0.7, 99.97)
 (0.8, 99.98)
 (0.9, 99.97)};

\nextgroupplot[xtick={0, 0.2, 0.4, 0.6, 0.8, 1.0}, xticklabel style={rotate=45, anchor=east}, ylabel=AUPR, xlabel=Margin]

\addplot[mark=x,blue]
    coordinates{
(0.0, 99.99)
 (0.1, 99.97)
 (0.2, 99.99)
 (0.3, 99.99)
 (0.4, 99.99)
 (0.5, 99.99)
 (0.6, 100.0)
 (0.7, 99.99)
 (0.8, 100.0)
 (0.9, 99.99)};
\addplot[mark=x,black]
    coordinates{
(0.0, 99.93)
 (0.1, 99.93)
 (0.2, 99.96)
 (0.3, 99.96)
 (0.4, 99.97)
 (0.5, 99.97)
 (0.6, 99.97)
 (0.7, 99.97)
 (0.8, 99.98)
 (0.9, 99.97)};


\nextgroupplot[xtick={0, 0.2, 0.4, 0.6, 0.8, 1.0}, xticklabel style={rotate=45, anchor=east}, ytick = {0, 0.1}, ylabel=FPR95, xlabel=Margin]

\addplot[mark=x,blue]
    coordinates{
(0.0, 0.0)
 (0.1, 0.02)
 (0.2, 0.0)
 (0.3, 0.0)
 (0.4, 0.0)
 (0.5, 0.0)
 (0.6, 0.0)
 (0.7, 0.0)
 (0.8, 0.0)
 (0.9, 0.0)};
\addplot[mark=x,black]
    coordinates{
(0.0, 0.13)
 (0.1, 0.12)
 (0.2, 0.05)
 (0.3, 0.04)
 (0.4, 0.02)
 (0.5, 0.03)
 (0.6, 0.02)
 (0.7, 0.01)
 (0.8, 0.0)
 (0.9, 0.0)};

\nextgroupplot[xtick={0, 0.2, 0.4, 0.6, 0.8, 1.0}, xticklabel style={rotate=45, anchor=east}, ylabel=AUROC, xlabel=Margin]

\addplot[mark=x,blue]
    coordinates{
(0.0, 87.89)
 (0.1, 89.06)
 (0.2, 89.06)
 (0.3, 88.69)
 (0.4, 87.84)
 (0.5, 87.6)
 (0.6, 90.45)
 (0.7, 85.56)
 (0.8, 86.77)
 (0.9, 87.38)};
\addplot[mark=x,black]
    coordinates {
(0.0, 89.27)
 (0.1, 91.16)
 (0.2, 92.25)
 (0.3, 92.11)
 (0.4, 92.45)
 (0.5, 91.49)
 (0.6, 91.53)
 (0.7, 88.66)
 (0.8, 91.15)
 (0.9, 90.63)};

\nextgroupplot[xtick={0, 0.2, 0.4, 0.6, 0.8, 1.0}, xticklabel style={rotate=45, anchor=east}, ylabel=AUPR, xlabel=Margin]

\addplot[mark=x,blue]
    coordinates{
(0.0, 86.41)
 (0.1, 87.67)
 (0.2, 87.47)
 (0.3, 87.35)
 (0.4, 86.46)
 (0.5, 86.25)
 (0.6, 89.15)
 (0.7, 83.54)
 (0.8, 85.81)
 (0.9, 86.18)};
\addplot[mark=x,black]
    coordinates{
(0.0, 87.72)
 (0.1, 90.09)
 (0.2, 91.04)
 (0.3, 90.94)
 (0.4, 91.27)
 (0.5, 90.47)
 (0.6, 90.28)
 (0.7, 87.32)
 (0.8, 89.93)
 (0.9, 89.14)};


\nextgroupplot[xtick={0, 0.2, 0.4, 0.6, 0.8, 1.0}, xticklabel style={rotate=45, anchor=east}, ylabel=FPR95, xlabel=Margin, legend to name=test2legend]

\addplot[mark=x,blue]
    coordinates{
(0.0, 35.16)
 (0.1, 34.23)
 (0.2, 31.83)
 (0.3, 32.53)
 (0.4, 33.79)
 (0.5, 34.48)
 (0.6, 29.55)
 (0.7, 41.05)
 (0.8, 34.82)
 (0.9, 35.43)};
\addplot[mark=x,black]
    coordinates{
(0.0, 36.95)
 (0.1, 33.81)
 (0.2, 29.02)
 (0.3, 28.74)
 (0.4, 27.76)
 (0.5, 30.66)
 (0.6, 29.79)
 (0.7, 35.68)
 (0.8, 30.5)
 (0.9, 29.11)};

\addlegendentry{Allconv}
\addlegendentry{WRN}

\end{groupplot}
\end{tikzpicture}
\ref{test2legend}
\caption{Line graph representing the OOD detection performance of MaCS across different margin values. Each row represents different ID datasets in the order from top to bottom: CIFAR-10, CIFAR-100, SVHN, Imagenet-32. The results represent an average value over multiple OOD datasets.}
\label{fig:figure_4}
\end{center}
\end{figure}
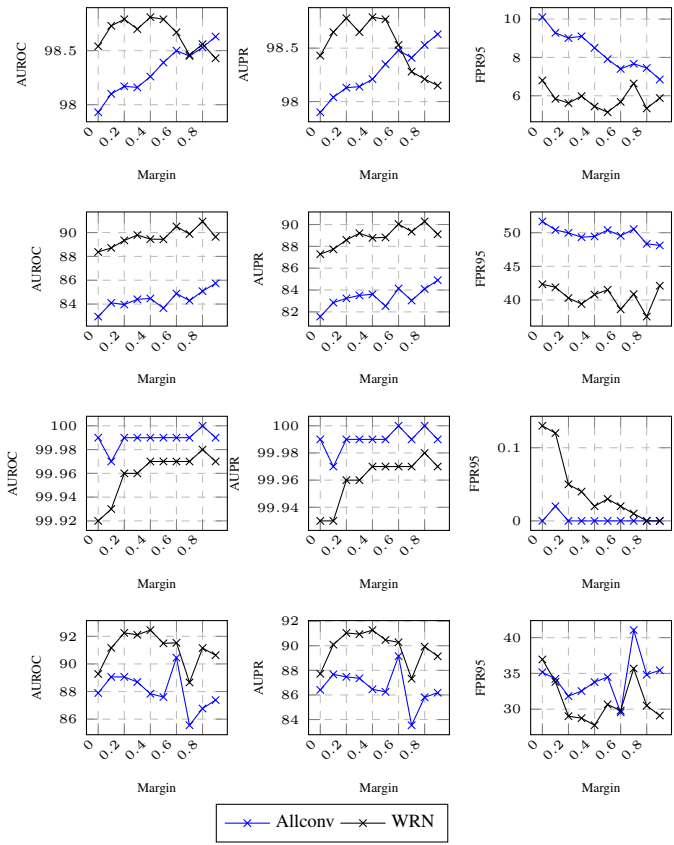

In this ablation study, we evaluated the detection performance of the proposed method under different values of $m$. We used a margin value $m\in\left\{ 0.0, 0.1, \hdots, 0.9\right\}$, and fine-tuned two models WRN and Allconv using all four ID datasets as mentioned in Section. \ref{sec:id_dataset}. Fig. \ref{fig:figure_4}, depicts the AUROC, AUPR, and FPR95 scores averaged over five different test OOD datasets against the range of values of $m$. From the figure, we can observe that the characteristics of the curve remains different for different ID datasets, nonetheless, for a particular ID dataset both models (WRN, and Allconv) exhibit similar trend throughout the values of $m$. Overall, the model is seen to perform best at or after $m=0.5$. In terms of the impact of $m$, most of the time larger values are expected to increase the dispersion of OOD and ID scores towards their respective likelihood limits of 0 and 1. We record the optimum detection results for each dataset, across both models and report it in Table. \ref{tab:table_5}. These results emphasize the importance of carefully selecting the value of $m$ to achieve optimal performance for MaCS.

\begin{table}[ht!]
    \caption{Optimum value of $m$ reported while using different ID data and models.}
    \centering
    \resizebox{0.45\textwidth}{!}{
    \begin{tabular}{|c|c|c|c|c|}
        \hline
         \backslashbox{\bf ID Data}{\bf Models} &  \bf WRN & \bf Allconv & \bf Resnet-18 & \bf Densenet-121\\
         \hline 
         \bf CIFAR-10 & 0.5 & 0.9 & 0.8 & 0.7\\ 
         \hline
         \bf CIFAR-100 &  0.8 & 0.9 & 0.8 & 0.8 \\
         \hline
         \bf SVHN & 0.8 & 0.8 & 0.8 & 0.9  \\
         \hline
         \bf Imagenet-32 & 0.5 & 0.6 & 0.5 & 0.6 \\
         \hline
    \end{tabular}}
    \label{tab:table_5}
\end{table}

\subsection{Training with Different Neural Networks}
After achieving favorable outcomes of MaCS and MaCS$^*$ with WRN, and Allconv, we sought to determine if this performance could be replicated on other models. To this end, we trained each of the reference methods, as well as MaCS, under similar training configurations, but using different backbone architectures, namely Resnet-18 and Densenet-121, both of which are widely used image classification models. We utilized all four ID datasets and all five outlier datasets. We report the test results in Tables \ref{tab:table_6}, \ref{tab:table_7} and it is evident that our methods consistently achieved the best or second-best performance across the majority of the test datasets. These findings confirm that our method's performance is not limited to a particular type of neural network, as it demonstrates the capacity to achieve optimal results regardless of the network employed.

\begin{table*}[ht!]
    \caption{Comprehensive OOD detection results obtained by training different ID datasets on different backbone architectures. Best, and second best results are represented in bold and underline respectively. For same results with other methods, we choose ours to be best or the second best.}
    \centering
    \begin{subtable}{\textwidth}
    \resizebox{\textwidth}{!}{
    \begin{tabular}{|l|c|ccc|ccc|}
        \hline
         \bf Method & \bf Models & \multicolumn{3}{c|}{\bf CIFAR-10} &\multicolumn{3}{c|}{\bf CIFAR-100} \\
         && \bf AUROC $\uparrow$ & \bf AUPR $\uparrow$ & \bf FPR95 $\downarrow$ &\bf AUROC $\uparrow$ & \bf AUPR $\uparrow$ & \bf FPR95 $\uparrow$ \\ 
         \hline
         \multirow{3}{3em}{OE} \cite{hendrycks2018deep} & Allconv & 97.90±0.03 & 97.86±0.04 & 10.18±0.19 & 83.07±0.12 & 81.83±0.18 & 52.11±0.18 \\
         
         & Resnet-18 & \underline{97.51±0.04} & \bf{97.37±0.06} & \underline{11.96±0.16} & 86.39±0.15 & 84.56±0.19 & 47.35±0.40\\
         
         & Densenet-121 & 96.71±0.06 & 96.29±0.10& 14.86±0.25 & 83.95±0.19 & 81.42±0.31 & \underline{52.68±0.38} \\
         \hline
         \multirow{3}{3em}{Energy} \cite{liu2020energy} & Allconv & 96.79±0.04 & 96.60±0.06 & 13.94±0.16 & 79.17±0.13 & 78.13±0.18 & 58.56±0.36\\
         
         & Resnet-18 & 97.46±0.06 & \underline{97.31±0.1} & 12.74±0.27 & 85.27±0.12 & 85.35±0.15 & 55.37±0.26 \\
         
         & Densenet-121 & 96.89±0.05 & \underline{96.70±0.08} & \underline{14.59±0.26} & 82.16±0.12 & 81.81±0.20& 63.99±0.33\\
         \hline
         
         \multirow{3}{3em}{MixOE} \cite{zhang2023mixture} & Allconv & 93.52±0.04 & 91.77±0.08 & 22.83±0.22 & 77.89±0.11 & 74.39±0.15 & 60.10±0.33  \\
         
         & Resnet-18 & 84.95±0.13 & 81.51±0.18 & 48.40±0.35 & 77.30±0.22 & 72.54±0.33 & 61.74±.38\\
         
         & Densenet-121 & 85.10±0.12 & 83.87±.15 & 57.69±0.38 & 74.18±.1 & 71.43±0.15 & 72.35±0.25 \\
         
         \hline
         \multirow{3}{3em}{DivOE} \cite{zhu2024diversified} & Allconv & 97.70±0.03 & 97.65±0.04 & 11.24±0.13 & 81.96±0.12 & 81.10±0.13 & 54.30±0.27  \\
         
         & Resnet-18 & 97.12±0.06 & 96.93±0.09 & 13.64±0.18 & 85.25±0.12 & 83.30±0.20& 50.24±0.29 \\
         
         & Densenet-121 & 96.33±0.06 & 95.98±0.11 & 16.58±0.30& \underline{84.00±0.12} & \underline{82.02±0.18} & 53.76±0.25 \\
         
         \hline
         \multirow{3}{3em}{MaCS} & Allconv & \underline{98.39±0.03} & \underline{98.35±0.04} & \underline{7.91±0.1} & \underline{83.66±0.13} & \underline{82.53±0.2} & \underline{50.39±0.31}\\
         
         & Resnet-18 & 97.00±0.05 & 96.56±0.10& 13.29±0.22 & \underline{87.39±0.13} & \bf{86.27±0.16} & \underline{47.12±0.27} \\
         
         & Densenet-121 & 95.99±0.05 & 95.59±0.08 & 18.35±0.30& 83.31±0.10& 81.75±0.16 & 56.88±0.19 \\
         
         \hline
         \multirow{3}{3em}{MaCS{$^*$}} & Allconv & \bf{98.63±0.04} & \bf{98.63±0.04} & \bf{6.85±0.18} & \bf{85.74±0.12} & \bf{84.90±0.09} & \bf{48.11±0.51} \\
         
         & Resnet-18 & \bf{97.61±0.03} & 97.28±0.05 & \bf{10.69±0.13} & \bf{87.47±0.1} & \underline{85.43±0.13} & \bf{44.80±0.27} \\
         
         & Densenet-121 & \bf{97.58±0.04} & \bf{97.33±0.08} & \bf{11.37±0.16} & \bf{85.37±0.12} & \bf{82.22±0.19} & \bf{49.89±0.27} \\
         \hline
    \end{tabular}}
    \caption{CIFAR-10 and CIFAR-100}
    \label{tab:table_6}
    \end{subtable}
    \vspace{2em} 
    \begin{subtable}{\textwidth}
        \centering
    \resizebox{\textwidth}{!}{
    \begin{tabular}{|l|c|ccc|ccc|}
        \hline
         \bf Method & \bf Models & \multicolumn{3}{c|}{\bf SVHN} & \multicolumn{3}{c|}{\bf Imagenet-32} \\
         && \bf AUROC $\uparrow$ & \bf AUPR $\uparrow$ & \bf FPR95 $\downarrow$ &\bf AUROC $\uparrow$ & \bf AUPR $\uparrow$ & \bf FPR95 $\uparrow$ \\ 
         \hline
         \multirow{3}{3em}{OE} \cite{hendrycks2018deep} & Allconv & 99.99±0.00& 99.99±0.0& 0.00±0.0& 87.63±0.03 & 86.13±0.03 &  36.14±0.06 \\
         
         & Resnet-18 & 99.99±0.00 & 99.99±0.00 & 0.00±0.00 & 92.05±0.03 & 91.01±0.04 &  \underline{27.79±.04} \\
         
         & Densenet-121 & 99.99±0.00 & 99.99±0.00 & 0.00±0.00 & 91.19±0.02 & 90.37±0.02 & 32.39±0.09  \\
         
         \hline
         \multirow{3}{3em}{Energy} \cite{liu2020energy} & Allconv & 99.99±0.00 & 99.99±0.0& 0.00±0.0& 87.06±0.03 & 85.95±0.02 &  37.56±0.08 \\
         
         & Resnet-18 & 99.97±0.00 & 99.97±0.00 & 0.01±0.00 & 90.23±0.03 & 89.30±0.03 & 31.11±0.06\\
         
         & Densenet-121 & 99.99±0.00 & 99.99±0.00 & 0.00±0.00 & 92.77±0.02 & 92.27±0.02 & 30.89±0.08 \\
         
         \hline
         \multirow{3}{3em}{MixOE} \cite{zhang2023mixture} & Allconv & 95.91±0.04 & 94.89±0.06& 15.03±0.13& 73.75±0.05 & 68.14±0.05 &  58.09±0.08 \\
         
         & Resnet-18 & 92.07±0.08 & 88.80±0.12 & 24.21±0.16 & 77.98±0.04 & 69.96±0.08 & 55.45±0.08\\
         
         & Densenet-121& 94.99±0.05 & 93.71±0.06 & 18.49±0.13 & 75.10±0.06 & 67.30±0.10& 63.11±0.1\\
         
         \hline
         \multirow{3}{3em}{DivOE} \cite{zhu2024diversified} & Allconv & 99.99±0.00 & 99.99±0.0& 0.00±0.0& \underline{87.97±0.02} & \underline{86.70±0.02} &  34.66±0.03 \\
         
         & Resnet-18 & \bf{100.00±0.00} & \bf{100.00±0.00} & \bf{0.00±0.00} & \underline{92.61±0.02} & \underline{91.65±0.03} & \bf{27.34±0.06}\\
         
         & Densenet-121 & \bf{100.00±0.00} & \bf{100.00±0.00} & \bf{0.00±0.00} & 91.35±0.02 & 90.59±0.03 & 31.47±0.08\\
         
         \hline
         \multirow{3}{3em}{MaCS} & Allconv & \underline{99.99±0.00} & \underline{99.99±0.00}& \underline{0.00±0.00}& 87.60±0.02 & 86.25±0.02 & \underline{34.48±0.06} \\
         
         & Resnet-18 & 99.99±0.00 & 99.99±0.00 & 0.00±0.00 & 92.81±0.02 &92.29±0.02 & 29.13±0.06\\
         
         & Densenet-121 & 99.99±0.00 & 99.99±0.00 & 0.00±0.00 & \underline{93.26±0.03} & \underline{92.47±0.03} & \underline{27.00±0.08}\\
         
         \hline
         \multirow{3}{3em}{MaCS{$^*$}} & Allconv & \bf{100.00±0.00} & \bf{100.00±0.00}& \bf{0.00±0.00}& \bf{90.45±0.02} & \bf{89.15±0.04} &  \bf{29.55±0.06} \\
         
         & Resnet-18 & \underline{99.99±0.00} & \underline{99.99±0.00} & \underline{0.00±0.00} & \bf{92.81±0.02} &\bf{92.29±0.02} & 29.13±0.06 \\
         
         & Densenet-121 & \underline{99.99±0.00} & \underline{99.99±0.00} & \underline{0.00±0.00} & \bf{94.40±0.2} & \bf{92.84±0.02} & \bf{25.94±0.06}\\
         
         \hline
    \end{tabular}}
    \caption{SVHN and Imagenet-32}
    \label{tab:table_7}
    \end{subtable}
    
    \label{tab:main_3}
\end{table*}

\subsection{Detection Performance with and without Margin bound}
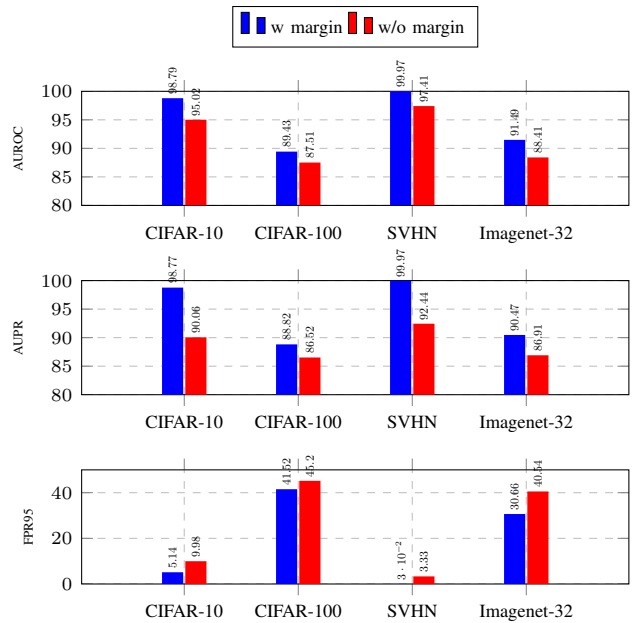
\begin{figure}
\centering

\begin{tikzpicture}
    
\begin{groupplot}
    [
    group style={
        group size=1 by 3,
        horizontal sep=1.08cm},
    ybar=0.8pt,
    symbolic x coords={CIFAR-10, CIFAR-100, SVHN, Imagenet-32},
    xtick=data,
    enlarge x limits=0.3, 
    legend cell align=left,
    width=\linewidth,
    height = 0.35\linewidth,
    legend style={at={(.5,4.7)},anchor=south, legend columns = 2}, grid=both, grid style=dashed, tick label style={font=\scriptsize}, label style={font=\tiny},legend style={font=\scriptsize},]
\nextgroupplot[ymin = 80, ymax = 100, ylabel = AUROC]
\addplot [ybar, bar width = 8pt, draw=none, fill=blue, nodes near coords, nodes near coords align = {vertical}, nodes near coords style={scale=0.45, rotate= 90, anchor=west}] coordinates {(CIFAR-10, 98.79) (CIFAR-100, 89.43) (SVHN, 99.97) (Imagenet-32, 91.49) };

\addplot [ybar, bar width = 8pt, draw=none, fill=red, nodes near coords, nodes near coords align = {vertical}, nodes near coords style={scale=0.45, rotate= 90, anchor=west}] coordinates {(CIFAR-10, 95.02) (CIFAR-100, 87.51) (SVHN, 97.41) (Imagenet-32, 88.41) };

\nextgroupplot[ymin = 80, ymax = 100, ylabel = AUPR]

\addplot [ybar, bar width = 8pt, draw=none, fill=blue, nodes near coords, nodes near coords align = {vertical}, nodes near coords style={scale=0.45, rotate= 90, anchor=west}] coordinates {(CIFAR-10, 98.77) (CIFAR-100, 88.82) (SVHN, 99.97) (Imagenet-32, 90.47) };

\addplot [ybar, bar width = 8pt, draw=none, fill=red, nodes near coords, nodes near coords align = {vertical}, nodes near coords style={scale=0.45, rotate= 90, anchor=west}] coordinates {(CIFAR-10, 90.06) (CIFAR-100, 86.52) (SVHN, 92.44) (Imagenet-32, 86.91) };

\nextgroupplot[ymin = 0, ymax = 50, ylabel = FPR95]

\addplot [ybar, bar width = 8pt, draw=none, fill=blue, nodes near coords, nodes near coords align = {vertical}, nodes near coords style={scale=0.45, rotate= 90, anchor=west}] coordinates {(CIFAR-10, 5.14) (CIFAR-100, 41.52) (SVHN, 0.03) (Imagenet-32, 30.66) };

\addplot [ybar, bar width = 8pt, draw=none, fill=red,nodes near coords, nodes near coords align = {vertical}, nodes near coords style={scale=0.45, rotate= 90, anchor=west}] coordinates {(CIFAR-10, 9.98) (CIFAR-100, 45.2) (SVHN, 3.33) (Imagenet-32, 40.54) };

\legend{w margin, w/o margin}

\end{groupplot}

\end{tikzpicture}
\caption{Bar-graph representing different OOD metrics for individual ID datasets with and without margin bound. A WRN model was trained on these ID datasets. }
\label{fig:figure_5}
\end{figure}

In this ablation study, we eliminated the bounded margin and relied solely on MCD to check the influence of $m$ to the overall performance. We trained a WRN backbone on all four ID datasets, and tested on all OOD datasets. The result is depicted in Fig. \ref{fig:figure_5} as a bar chart, where we can observe a significant decline in the performance across all ID datasets when MaCS is not subjected to a margin bound. Although MCD assigns a penalty of zero to score explosions, it is evident that these values remain ambiguous and do not contribute to learning when not substituted with a specific weight, which is the value of $m$ in this instance. In essence, when one considers \eqref{eq:4}, and when margin is not used, $\mathcal{W}_{MaCS}$ will either assume a value of 0 or simply the MCD value, which may be null or the difference between ID and OOD scores. However, this difference does not correspond to the desired difference that is obtainable with margin.

\section{Conclusion}
In this paper, we introduced a novel and straightforward methodology aimed at improving OOD detection by establishing a compact decision boundary between ID and OOD data. To this end, we recognized a disguised OOD detection problem that existed in OE setting, i.e., score explosions, and proposed a solution, MaCS which first penalizes score explosions, and then substitutes it with a margin value to realize the difference between ID and OOD data to be as large as possible. Our approach significantly enhanced the OOD detection and provided competitive performance when compared with several S.O.T.A benchmarks across four ID datasets, and five OOD datasets in the image classification domain. Importantly, our proposed method was also able to achieve significant gain in ID accuracy. To summarize the detection performance, our method exhibited a remarkable gain in AUROC, AUPR, and FPR95, reaching a maximum improvement of 2.73\%, 3.26\%, and 9.06\%, respectively. These results affirm its effectiveness and thus demonstrate the synergy of OE with our method in advancing the field of OOD detection.

\subfour{Reproducibility: } All of our source codes, pretrained models, and results are publicly shared at: \url{https://github.com/lakpa-tamang9/margin_ood}

\bibliographystyle{ieeetr}
\bibliography{references}

\end{document}